\documentclass{article}

\usepackage{microtype}
\usepackage{graphicx}
\usepackage{booktabs} 
\usepackage{multirow}
\usepackage{amsmath}
\usepackage{caption}
\usepackage{subcaption}
\usepackage{adjustbox}
\usepackage{dblfloatfix}    

\usepackage{hyperref}

\usepackage[accepted]{mlsys2020}

\mlsystitlerunning{A Comprehensive Benchmark for Deep Learning Workload on MIG}

\begin{document}

\twocolumn[
\mlsystitle{MIGPerf: A Comprehensive Benchmark for Deep Learning Training and Inference Workloads on Multi-Instance GPUs}


\mlsyssetsymbol{equal}{*}

\begin{mlsysauthorlist}
\mlsysauthor{Huaizheng Zhang}{to}
\mlsysauthor{Yuanming Li}{to}
\mlsysauthor{Wencong Xiao}{to}
\mlsysauthor{Yizheng Huang}{as}
\mlsysauthor{Xing Di}{wu}
\mlsysauthor{Jianxiong Yin}{nvidia}
\mlsysauthor{Simon See}{nvidia}
\mlsysauthor{Yong Luo}{wu}
\mlsysauthor{Chiew Tong Lau}{ntu}
\mlsysauthor{Yang You}{nus}
\end{mlsysauthorlist}

\mlsysaffiliation{to}{Alibaba Group}
\mlsysaffiliation{wu}{Wuhan University}
\mlsysaffiliation{as}{Institute for Infocomm Research, A*STAR}
\mlsysaffiliation{nvidia}{NVIDIA AI Tech Center}
\mlsysaffiliation{ntu}{Nanyang Technological University}
\mlsysaffiliation{nus}{National University of Singapore}

\mlsyscorrespondingauthor{Yuanming Li}{yuanming.li@alibaba-inc.com}

\mlsyskeywords{Deep Learning, MLSys}

\vskip 0.3in

\begin{abstract}
New architecture GPUs like A100 are now equipped with multi-instance GPU (MIG) technology, which allows the GPU to be partitioned into multiple small, isolated instances. This technology provides more flexibility for users to support both deep learning training and inference workloads, but efficiently utilizing it can still be challenging. The vision of this paper is to provide a more comprehensive and practical benchmark study for MIG in order to eliminate the need for tedious manual benchmarking and tuning efforts. To achieve this vision, the paper presents MIGPerf, an open-source tool that streamlines the benchmark study for MIG. Using MIGPerf, the authors conduct a series of experiments, including deep learning training and inference characterization on MIG, GPU sharing characterization, and framework compatibility with MIG. The results of these experiments provide new insights and guidance for users to effectively employ MIG, and lay the foundation for further research on the orchestration of hybrid training and inference workloads on MIGs. The code and results are released on https://github.com/MLSysOps/MIGProfiler. \textit{This work is still in progress and more results will be published soon.}

\end{abstract}
]



\printAffiliationsAndNotice{}  

\section{Introduction}
\label{introduction}

Deep learning (DL) models have been widely deployed in many industries, ranging from e-commerce \cite{liu2016deepfashion, wu2019m2e, zhang2020look} and video analysis \cite{liu2020deep, oprea2020review, zhang2020deepqoe, jiao2021new}, to autopilot \cite{grigorescu2020survey} and finance \cite{heaton2017deep, ozbayoglu2020deep}. To support these computation-intensive models efficiently, users have to use new hardware devices like GPUs. Among these GPUs, newly released GPUs like A100 and A30, equipped with multi-instance GPU (MIG) technology \cite{nvidiamig, choquette2021nvidia}, have attracted attention. With the help of MIG, a whole GPU like A100 can be partitioned into several isolated small GPU instances (GI), providing more flexibility to support DL training and inference workloads. For instance, users can partition an A100 into seven GIs (1/7 GI) to serve small-sized models like ResNet50 \cite{he2016deep}. By doing so, an A100 can support seven ResNet50 models parallelly, but still meet Service-Level-Objectives (SLOs). Also, when training large language models like 
RoBERTa \cite{liu2019roberta}, users can use the whole GPU (7/7 GI) with no partitions. Furthermore, users are able to set up three 4/7, 2/7, and 1/7 GIs, and perform both training (on 4/7 GI) and inference (on 2/7 and 1/7 GIs) workloads simultaneously. This flexibility offers new chances to improve GPU utilization and save costs.


Though this flexibility offers new chances to improve GPU utilization and save costs, it also poses challenges. First, NVIDIA limits the partition by setting up hard-coded rules \cite{tan2021serving}. For instance, users can not have both 4/7 and 3/7 GIs simultaneously for an A100. As a result, users are not very free to partition the GPUs the same as CPUs or disks. Second, as each partitioned GI is isolated and has its own capacity (e.g., L2 cache, memory, etc), understanding different DL workloads' performance on GIs requires lots of manual benchmarking work. Last, the above-mentioned two issues result in that current scheduling methods for allocating CPUs and memory can not be easily applied to clusters with new MIG GPUs.


These issues lead to more and more work studying MIG performance under different workloads. Two representative works take a big step towards utilizing MIG GPUs better. The first study \cite{kaas2022deep} focuses on DL training. It uses three image recognition models, ResNet-26 (small), ResNet-50 (medium), and ResNet-152 (large), to explore how MIG GPUs perform training workloads with different model sizes. The second work \cite{tan2021serving} designs inference serving on MIG GPUs. It first benchmarks dozens of models from PyTorch and Tensorflow hubs \cite{pytorchhub, tensorflowhub} under different batch sizes and MIG partitions. Then it designs a serving system including an optimizer and a controller to serve DL inference workloads more cost-efficiently (saving 40\% GPU cost).

Though these works shed light on utilizing MIG, we still lack a \textbf{comprehensive} and \textbf{practical} benchmark study for MIG. Specifically, this limitation is reflected in three aspects. First, previous studies focus only on one workload, either training or inference, limiting their scope. Thus we need a holistic benchmark study for both training and inference workloads. Second, many powerful GPUs without MIG capabilities still dominate the cluster. These GPUs can only use software-based sharing methods and a comparison between MIG and software-based GPU sharing in the real world is still missing. Third, the compatibility of current training and serving frameworks with MIG is unclear. Understanding this compatibility can help to add new features to current frameworks as well as benefit MIG optimization, and thus should be treated as a high priority. Last, existing studies lack many important metrics like energy consumption, which is vital for green AI \cite{schwartz2020green}.


To address these issues and facilitate the research in this domain, we propose to explore MIG from multiple views, so as to provide a comprehensive and practical benchmark report. First, it examines MIG in practice with many real-world training and inference workloads and fixes several potential issues during benchmarking. Second, we compare MIG with software-based GPU sharing methods under a variety of settings to understand their suitable scenarios. Third, DL frameworks' compatibility to MIG is also explored. 

Furthermore, we implement an open-source tool, MIGPerf, aiming to streamline the MIG benchmark process. It first abstracts a general workflow from benchmarking training and inference workload on MIG GPUs and can help to start benchmarking with a few steps. Second, it offers APIs with highly a modulized backend and can be easily extended. Specifically, MIGPerf includes a controller to partition the MIG GPUs into different GIs, and a MIG profiler to generate benchmark workload and collect results. The results will be formatted so that third-party tools like Jupyter and Prometheus can quickly utilize them. Moreover, the tool is implemented based on Python and integrates many tools like DCGM to monitor performance on clouds more easily.

\begin{figure}[t]
  \centering
  \includegraphics[width=0.8\linewidth]{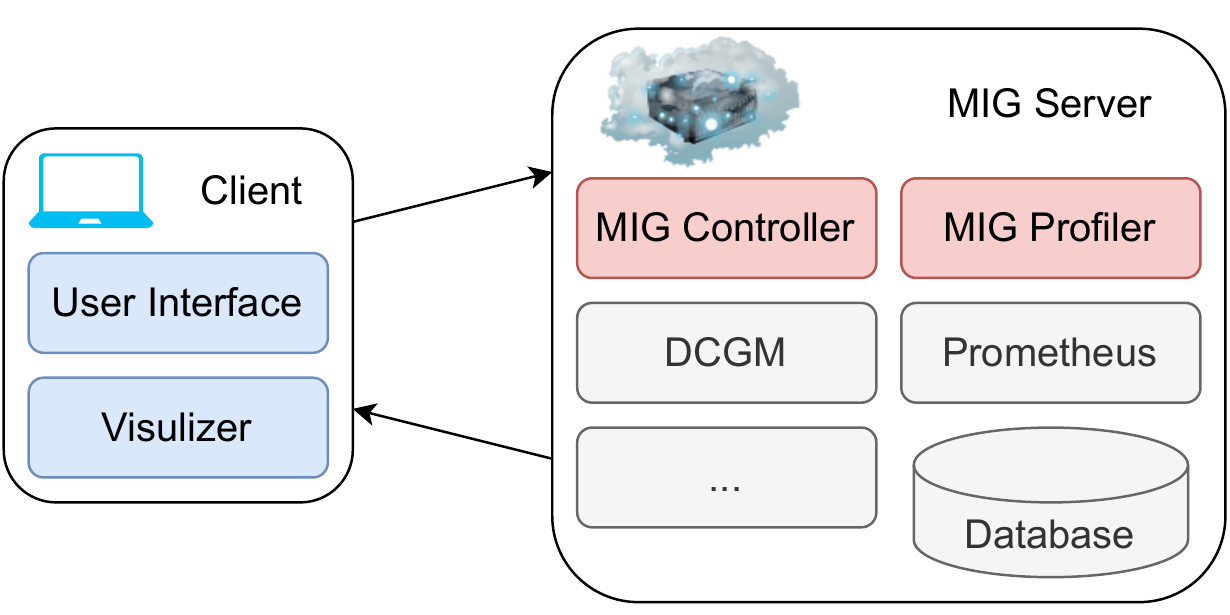}
  \caption{The overview of the proposed benchmarking system. The system first accepts users' benchmarking tasks. Then it distributes the tasks to dedicated servers to complete them automatically. Finally, it will send a detailed report and guidelines back to users.}
  \label{fig:benchmark_server}
\end{figure}

We employ MIGPerf to conduct many benchmarking studies on two MIG GPUs, A100 and A30, as a preliminary study. The smooth benchmarking process shows its effectiveness. Meanwhile, many new benchmarking results, ranging from sequence length impact and energy consumption to GPU sharing comparison and framework impact, are presented to reveal more insights. Besides, these preliminary studies show several very promising research directions. For instance, we may design a hybrid DL training and inference workload orchestration on MIG GPUs. Also, how to integrate the MIG into current serving frameworks to save inference costs remains an open problem. 


In the remainder of this paper, we first introduce the related work in Section \ref{sec:relatedwork}. Next, we present the system implementation and the employed methodologies in Section \ref{sec:system_design}. We employ our system to perform benchmark tasks and evaluate its performance in Section \ref{sec:result}. Finally, we summarize our paper and point out some future directions in Section \ref{sec:summary}.

\section{Related Work}
\label{sec:relatedwork}

We group the related work into two classes, deep learning (DL) benchmark, and GPU sharing. We briefly introduce them in this section.

\subsection{Deep Learning Benchmark}

Benchmark tools play a vital role in driving DL's development. These tools can be classified into two categories, macro-benchmark and micro-benchmark. Macro-benchmarks like DawnBench \cite{coleman2017dawnbench}, Fathom \cite{adolf2016fathom}, AI Benchmark \cite{ignatov2019ai} and MLPerf Inference \cite{reddi2020mlperf} are high-level models or hardware evaluations. These tools collect many models and run them on different hardware to explore their performance with many metrics like latency, throughput, etc. Micro-benchmarks explore the low-level components, which decide the speed of a model or hardware. Representative studies like AI matrix \cite{aimatrix} and DeepBench \cite{deepbench}  investigates how layers and computation kernels impact the final model performance.

\begin{figure*}[ht]
\begin{subfigure}{0.5\columnwidth}
  \centering
  \includegraphics[width=1.0\linewidth]{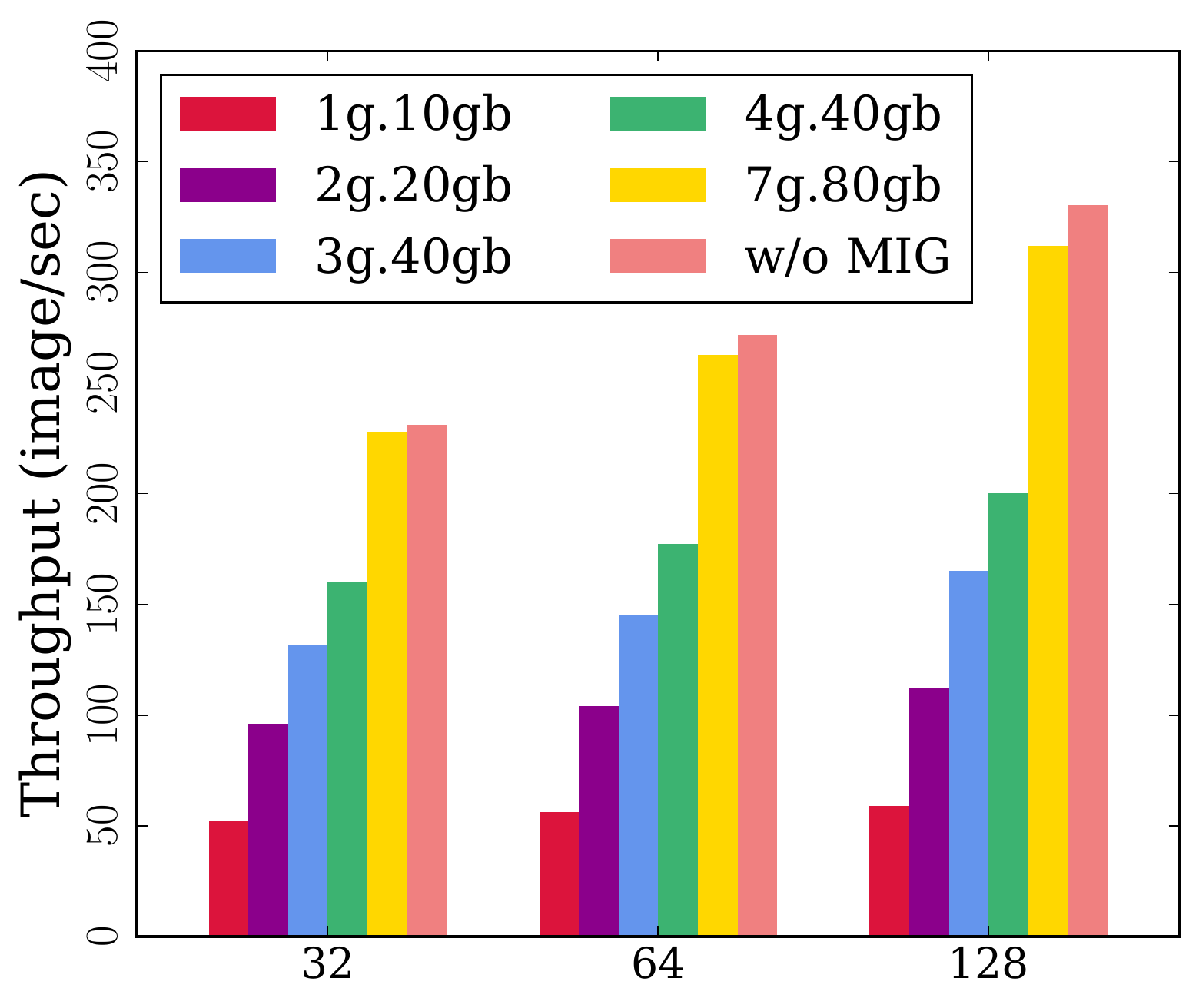}
  \caption{Throughput (batch/sec)}
  \label{fig:single_training_bert_qps}
\end{subfigure}%
\begin{subfigure}{.5\columnwidth}
  \centering
  \includegraphics[width=1.0\linewidth]{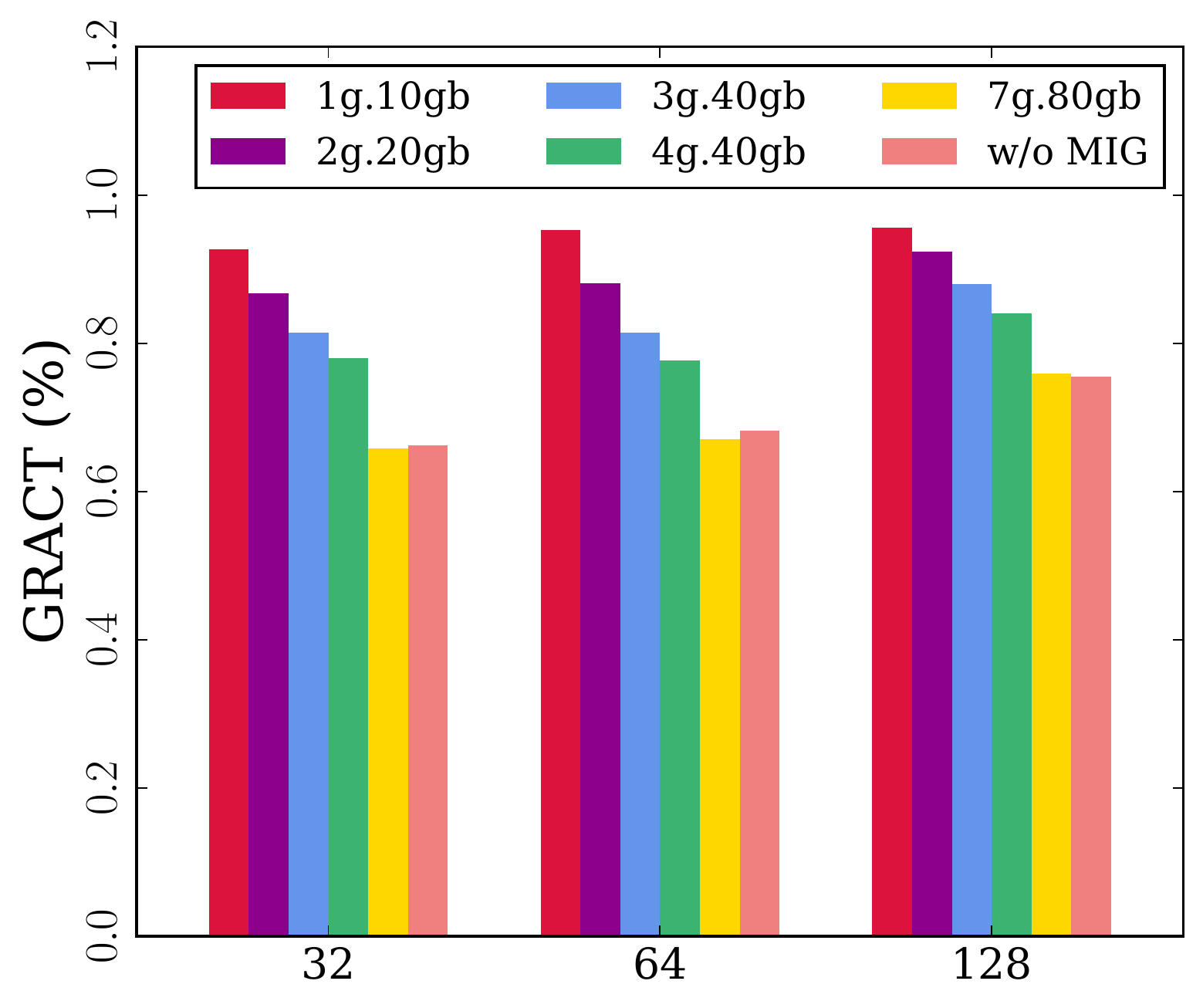}
  \caption{GRACT (\%)}
  \label{fig:single_training_bert_gract}
\end{subfigure}%
\begin{subfigure}{0.5\columnwidth}
  \centering
  \includegraphics[width=1.0\linewidth]{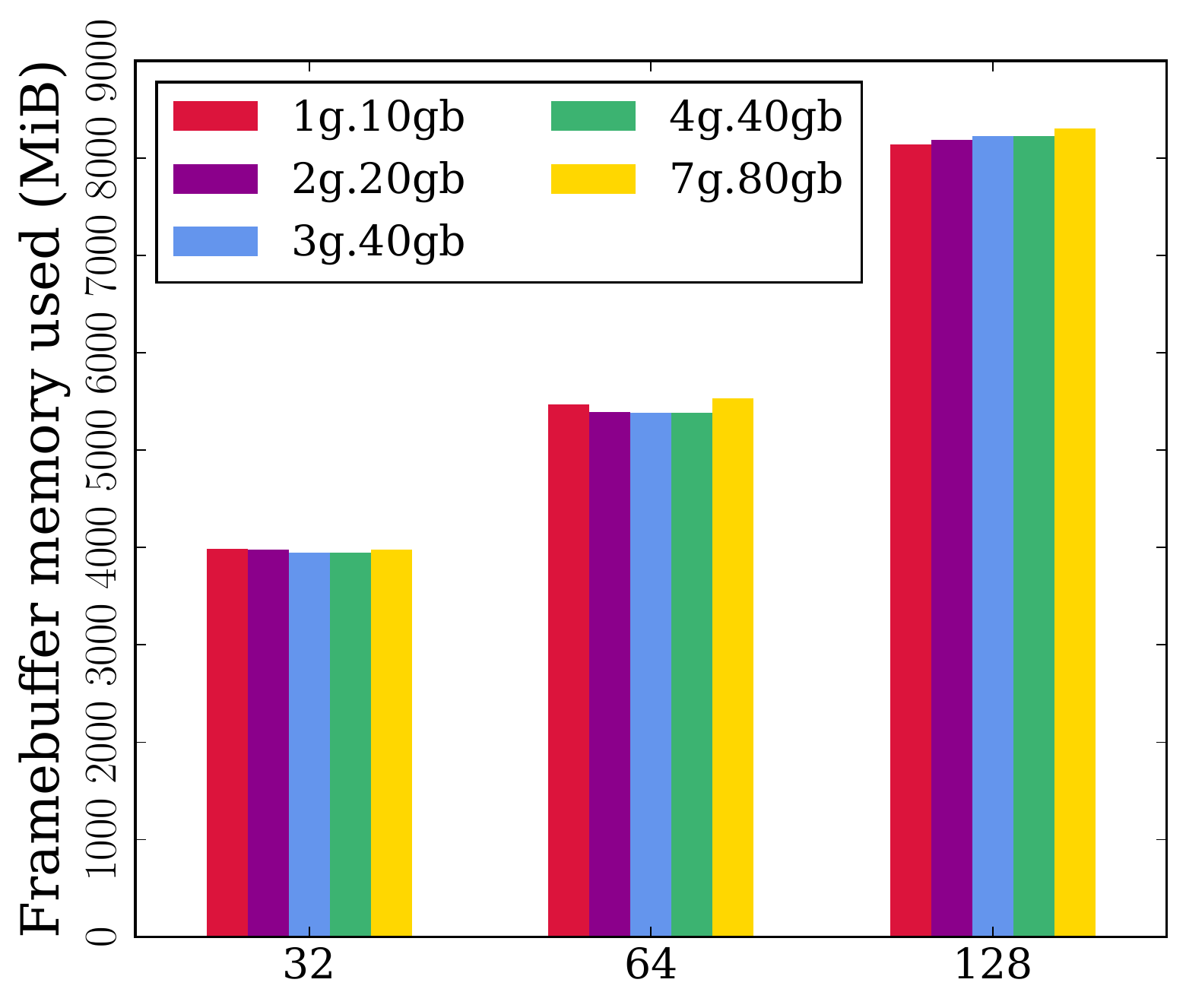}
  \caption{Framebuffer Memory (MiB)}
  \label{fig:single_training_bert_fb_memory}
\end{subfigure}%
\begin{subfigure}{.5\columnwidth}
  \centering
  \includegraphics[width=1.0\linewidth]{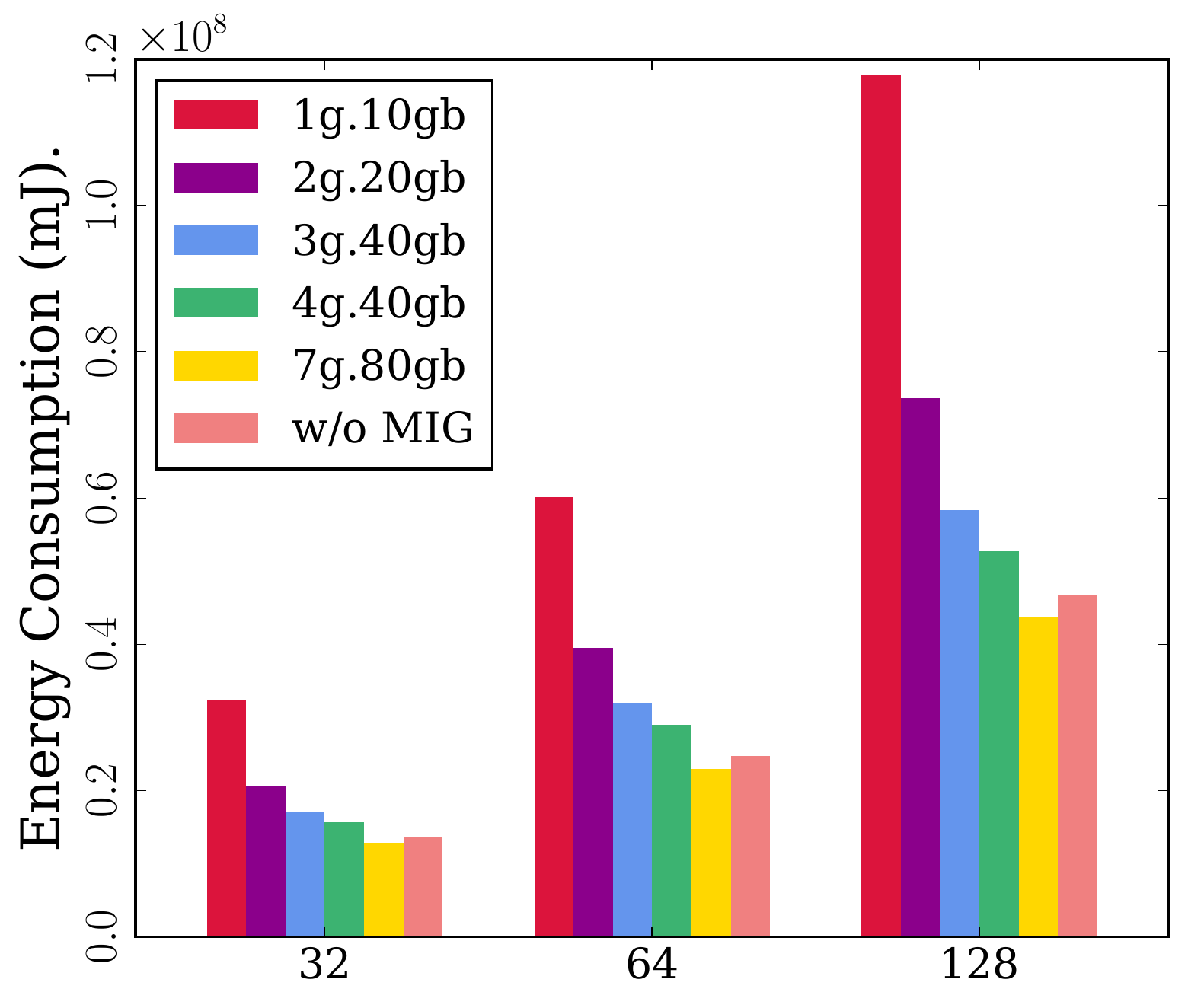}
  \caption{Energy Consumption (mJ)}
  \label{fig:single_training_bert_bs_energy}
\end{subfigure}
\caption{The impact of input batch size length for training on single GPU instance (GI) from A100. We present the batch size's influence on throughput, computation utilization, memory utilization, and energy consumption.}
\label{fig:single_training_bert_batch_size}
\end{figure*}

Two recent MIG studies can be classified into micro-benchmarks. One uses three ResNet models to explore the training performance with MIG, and the other study the inference performance for designing a better inference scheduler on MIG. In comparison, our work offers an open-source tool for users to explore both training and inference performance on MIG with ease. Meanwhile, more metrics, frameworks (e.g., Triton \cite{nvidiatritonserver}), and new models have been considered to provide a holistic performance evaluation of MIG. Besides, we provide a public leaderboard to continuously update the recent benchmark studies on MIG. In general, our system is a complement to existing benchmark tools.

\subsection{GPU Sharing}
To improve GPU utilization for DL workloads, recent studies have invested a lot of effort. Multi-Process Service (MPS) \cite{nvidiamps} technology from NVIDIA is designed to avoid costly context switches caused by multiple workloads in the same GPU. As a result, these workloads can achieve a relatively stable speed in a shared GPU while still meeting the service-level objective (SLO). However, as MPS can not provide real isolation, the interference can not be avoided especially for large workloads. To alleviate the issue, Gandiva \cite{xiao2018gandiva} proposes a fallback method for switching-sharing and non-sharing modes. Also, Salus \cite{yu2019salus} tries to build a primitive for fine-grained GPU sharing based on profiled DL workloads' memory usage. Moreover, many companies build virtual GPU (vGPU) \cite{nvidiavgpu} techniques so many cloud DL workloads can share these vGPUs according to their demands. Compared to them, MIG is the first technology provided by NVIDIA to achieve configurable physical isolation. However, we still lack a comprehensive and practical study to compare these techniques. This work provides some preliminary results and we will continue exploring this.

\section{System Design and Implementation}
\label{sec:system_design}

This section first presents the system workflow from the users' perspective. Then it introduces several core module implementations.

\begin{figure*}[t]
\begin{subfigure}{0.5\columnwidth}
  \centering
  \includegraphics[width=1.0\linewidth]{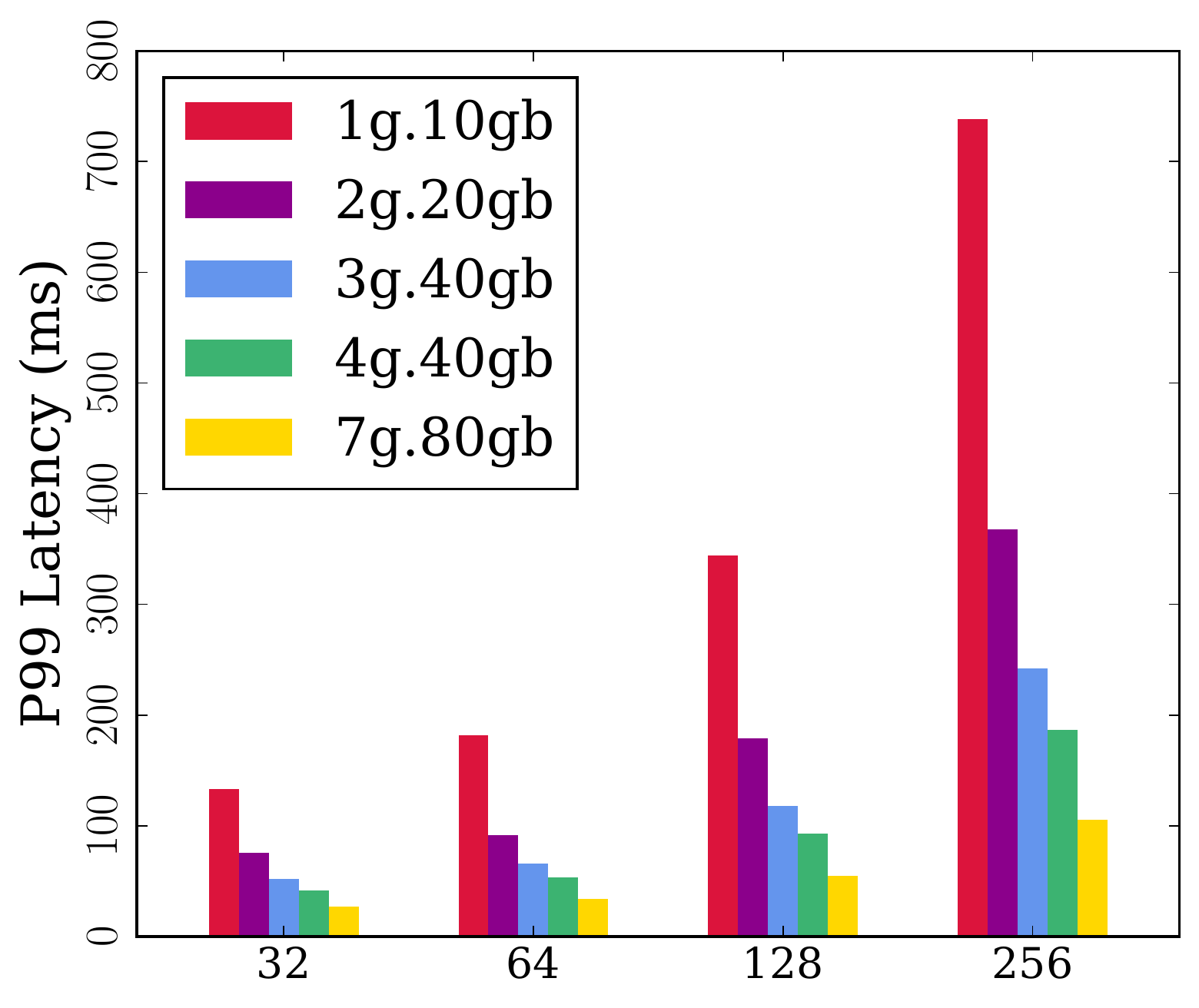}
  \caption{P99 Latency (ms)}
  \label{fig:single_inference_bert_p99_latency}
\end{subfigure}%
\begin{subfigure}{.5\columnwidth}
  \centering
  \includegraphics[width=1.0\linewidth]{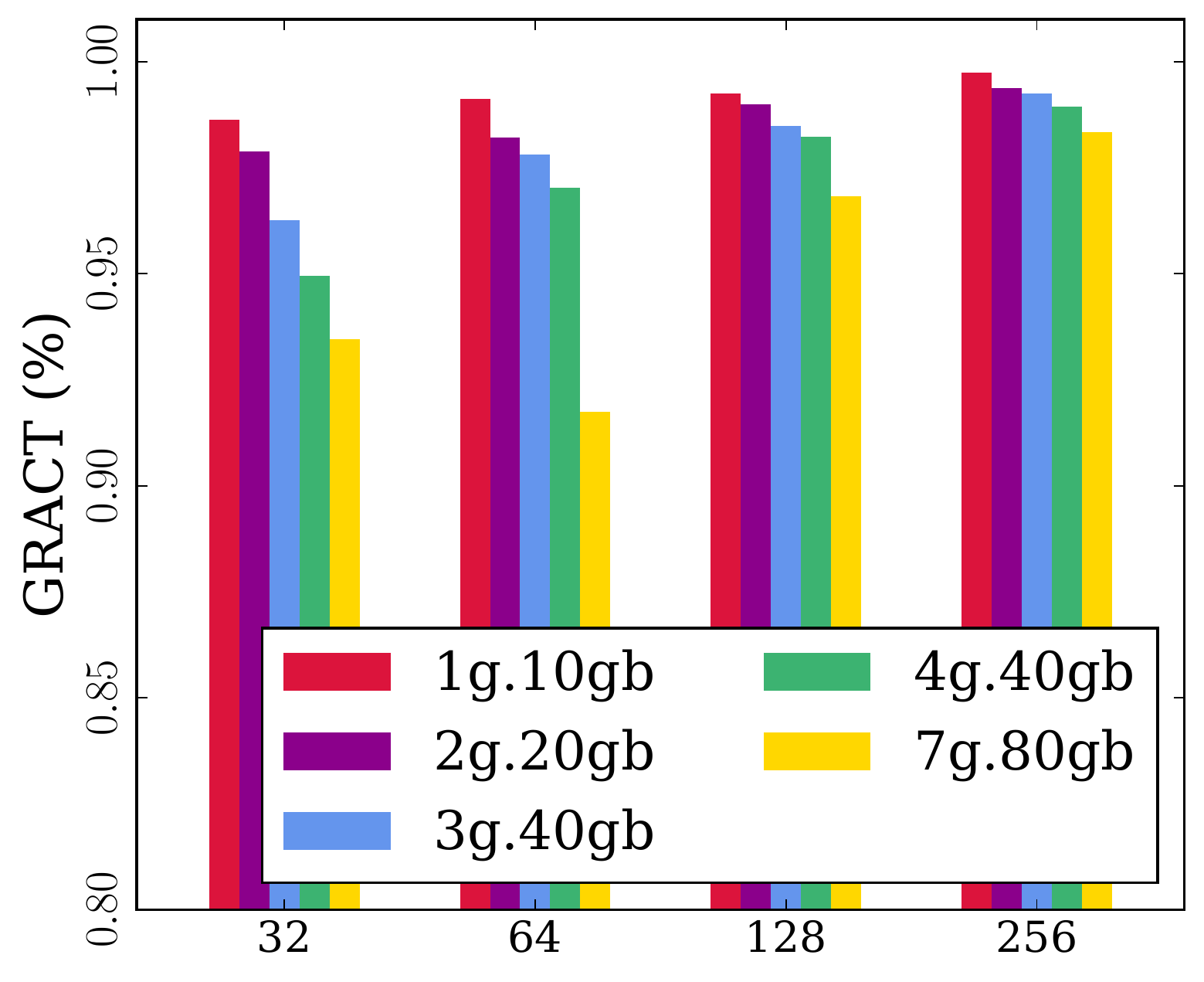}
  \caption{GRACT (\%)}
  \label{fig:single_inference_bert_utilization}
\end{subfigure}%
\begin{subfigure}{0.5\columnwidth}
  \centering
  \includegraphics[width=1.0\linewidth]{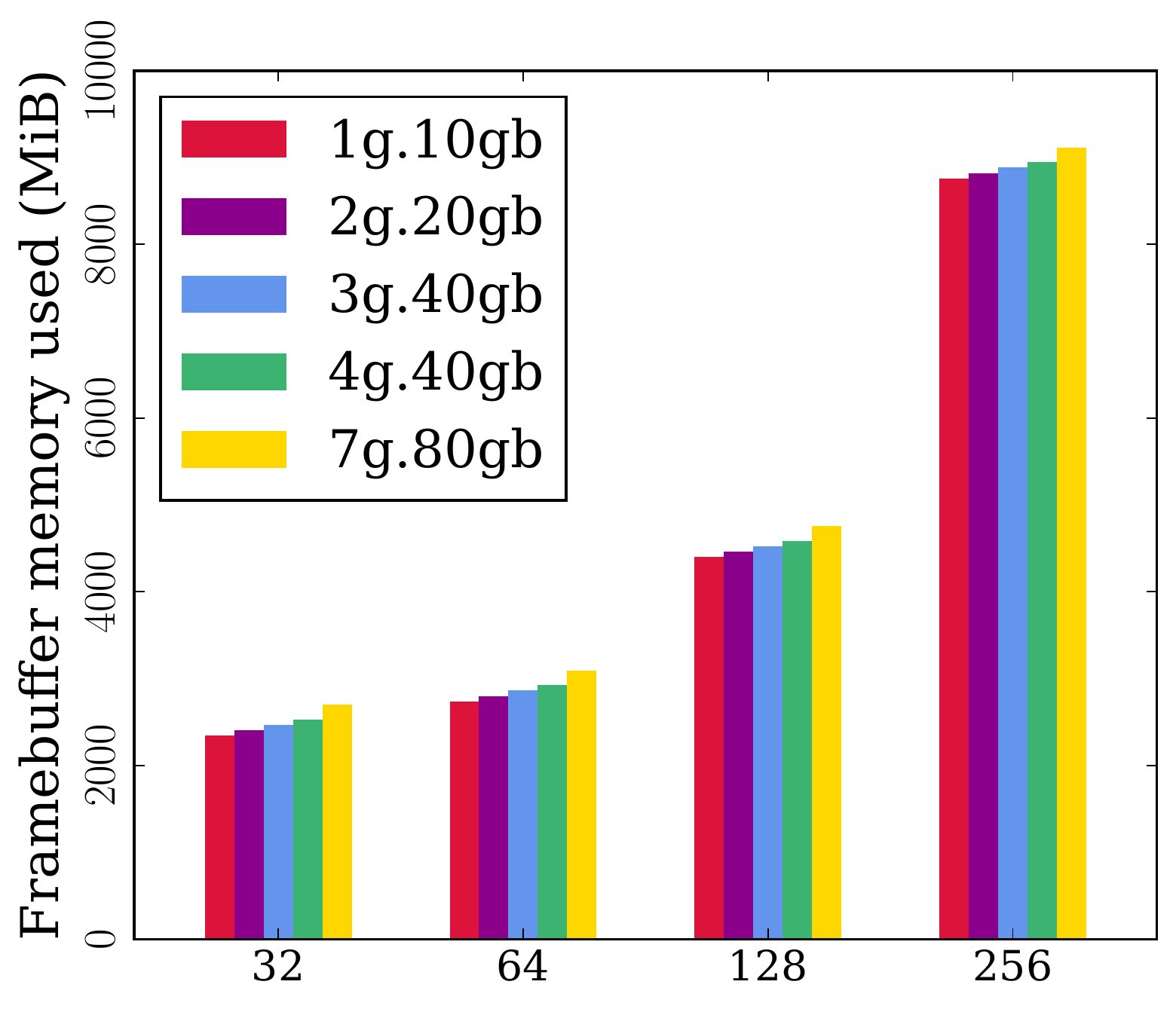}
  \caption{Framebuffer Memory (MiB)}
  \label{fig:single_training_bert_fb}
\end{subfigure}%
\begin{subfigure}{.5\columnwidth}
  \centering
  \includegraphics[width=1.0\linewidth]{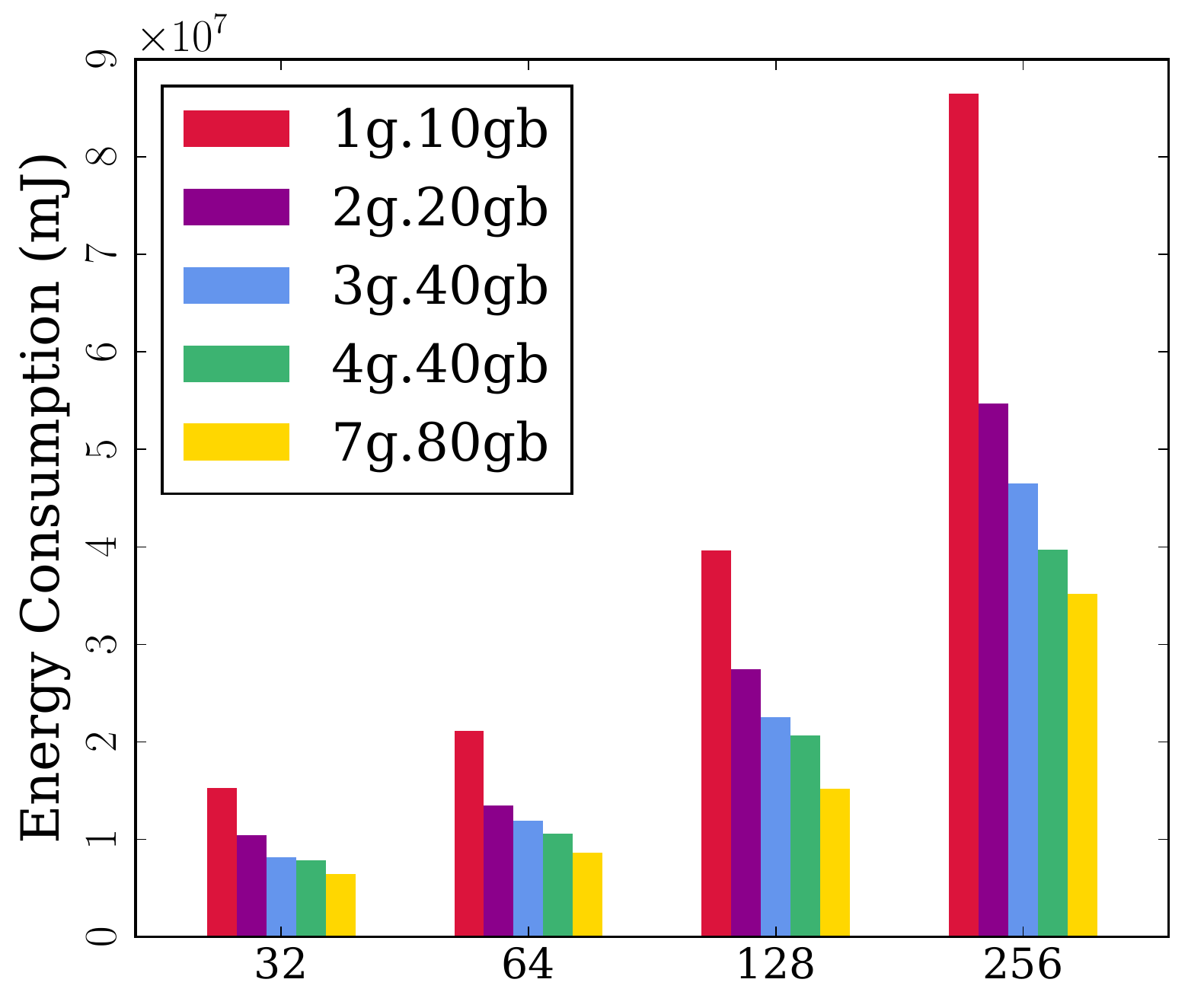}
  \caption{Energy Consumption (mJ)}
  \label{fig:single_inference_bert_sequence_energy}
\end{subfigure}
\caption{The impact of input sequence length for inference on single GPU instance (GI) from A100.}
\label{fig:single_inference_bert_sequence}
\end{figure*}

\subsection{System Overview}
\label{sec:system_overview}

In the beginning, our MIGPerf should be installed and started on a server with MIG GPUs.  Then users can call the API of the MIG Controller to partition GPUs into different GPU instances (GIs) according to their benchmark needs. Next, users can invoke the MIG Profiler by specifying the workload type (e.g., training), the model name, and the other settings (e.g., batch size) to perform a benchmark process. During the benchmark, the results will be saved into a local file in a time series manner. Meanwhile, the saved results will also be exported to different formats so that other well-deployed tools like Prometheus can directly consume them. Finally, users can employ the provided visualizer to quickly analyze the results for insights. Besides, users can install the client of MIGPerf on their own laptops to remotely control the whole process and conduct analysis locally.

\subsection{Core Components}
\label{sec:core_components}

\textbf{MIG Controller} helps users to manage the partitioned GIs in a more human-readable manner. It offers the python APIs to 1) enable MIG on a GPU, 2) operate the partition process, and 3) track the GIs. For each GI, the controller help to further build or destroy compute instances (CIs). By doing so, users will have more flexibility. For instance, the computation resources for jobs running in the same GI can be isolated while the memory resources can be shared. The module is built based on NVIDIA's MIG commands.

\textbf{MIG Profiler} abstracts the general deep learning (DL) training and inference workloads and monitors their running performance with many metrics (details are in A). It includes two parts, the workload performer and the performance aggregator. The workload performer simulates the training and inference processes and performs a benchmark according to user-specified settings. It is implemented atop many widely used model development frameworks like HuggingFace so users have little learning barrier to extend the module. The performance aggregator monitors the workload performance and system resource usage and saves them in the database. It is developed based on tools like DCGM.

\textbf{The other components} include a results exporter, a visualizer, and a client. All of them are designed to improve users' experience. For example, the exporter can format the saved performance results so they can be demonstrated with different performance analysis tools. The client can be installed on users' laptops for controlling the whole process remotely.

\section{Benchmark Study}
\label{sec:result}

This section presents several representative benchmark cases as preliminary studies. We first introduce our evaluation settings briefly and then detail the evaluation metrics. Finally, we discuss the experimental results on both training and inference workloads. Detailed settings and more results can be seen in Appendix.

\subsection{Evaluation Settings}
\label{eval_setting}

\textbf{Hardware}. We conduct experiments on two types of GPUs with MIG support, A100 and A30, respectively. The two GPUs are installed on two servers, separately, and the server details can be checked in Appendix \ref{sec:exp_detail}. 

\textbf{Models}. We select candidate benchmark models from highly recognized public model repositories like Hugging Face Transformers, TorchHub, etc. From these repositories, we choose models from a wide range of domains including image classification (e.g., vision transformer), language modeling (e.g., multilingual BERT), image generation (e.g., diffusion), etc. More model details are in Appendix \ref{sec:exp_detail}.

\subsection{Evaluation Metrics}
\label{eval_metric}

\textbf{Latency} measures the processing speed. We use two metrics, average latency and tail latency for inference workloads. Average latency measures the average processing time for a batch of requests, while tail latency means that the X-th (e.g. 99) percentile of requests has lower latency than the given value.

\textbf{Throughput} measures the training samples or inference requests that a device can process within a time unit (e.g., 1 min).

\textbf{Graphics Engine Activity (GRACT)} measures GPU utilization. It is the ratio of the allocated computation resources to the total available resources. Besides, it can be used to measure each partitioned GPU instance's utilization by utilizing the formulation provided by NVIDIA, which has been implemented in our tool.

\textbf{Frame Buffer (FB)} measures the occupied GPU memory of a running workload. Though this can be calculated by summing the model and input data sizes, we provide the results of FB to eliminate this manual effort.

\textbf{Energy consumption} is an estimation of electricity used for running a workload within a specific period of time(e.g., 5mins). We offer the results for users to understand if MIG can help to save electricity costs.

\subsection{MIG Training Characterization}
\label{sec:mig_training_exp}

The first experiment explores the MIG training performance under different MIG partition sizes and batch sizes. We present the BERT model's results, as shown in Figure \ref{fig:single_training_bert_batch_size} (More results can be checked in Appendix \ref{Appendix_results}.

Figure \ref{fig:single_training_bert_qps} shows that for small instances like 1g.10gb, increasing the batch size can not help to increase the throughput. We attribute the reason to the fully utilize computation resources of a GI. Since batch size = 32 already occupies all computation resources, increasing the batch size will lower the processing speed. Thus, the throughput will not be increased. Figure \ref{fig:single_training_bert_gract} also proves this. The utilization is very high and stable for small instances, whereas the large instances have a lower utilization value.

Figure \ref{fig:single_training_bert_fb_memory} presents the memory utilization. When the batch size is fixed, the memory usage has no difference across the GIs with different sizes. We own this to the large memory. Even for the smallest GIs, it can handle BERT easily.

Figure \ref{fig:single_training_bert_bs_energy} exhibits the energy consumption. It is no surprise that the small batch size will consume less energy. Surprisingly, under the same batch size, the larger the instance, the less energy it consumes. We guess the reason is faster processing speed for larger GIs. As we send a fixed number of requests, larger GIs will complete the processing more quickly, resulting in lower energy consumption.

\subsection{MIG Inference Characterization}
\label{sec:ming_inference_exp}

This section presents the inference benchmark on MIG. We use BERT as an example and more results are in Appendix \ref{Appendix_results}. Figure \ref{fig:single_inference_bert_p99_latency} presents the latency results. The latency is influenced by the batch size a lot when the GI partition size is small but this influence is marginal for the large GI. This can be attributed to the excessive computation resources of large GIs. Figure \ref{fig:single_inference_bert_utilization} shows the GPU utilization. For all batch sizes, the utilization is maintained in a high level. And as the GI size increase, the utilization becomes smaller, indicating workload can not utilize the GIs fully. Figure \ref{fig:single_training_bert_fb} presents the memory usage. The increase is marginal when the batch size is small but is large as the batch size increases. We think this is because the model size dominates the memory in the beginning and is then replaced by the large batch inputs. Figure \ref{fig:single_inference_bert_sequence_energy} shows the energy results which are the same as the training characterization.

\subsection{GPU Sharing Characterization}
\label{sec:mig_mps_exp}

In this section, we run multiple DL workloads on MIG-enabled or MPS-enabled machines, to explore the performance of these two GPU-sharing technologies. All of these experiments are conducted on an A30 GPU with PyTorch framework unless otherwise stated.

\begin{figure}[t]
\begin{subfigure}{0.5\columnwidth}
  \centering
  \includegraphics[width=1.0\linewidth]{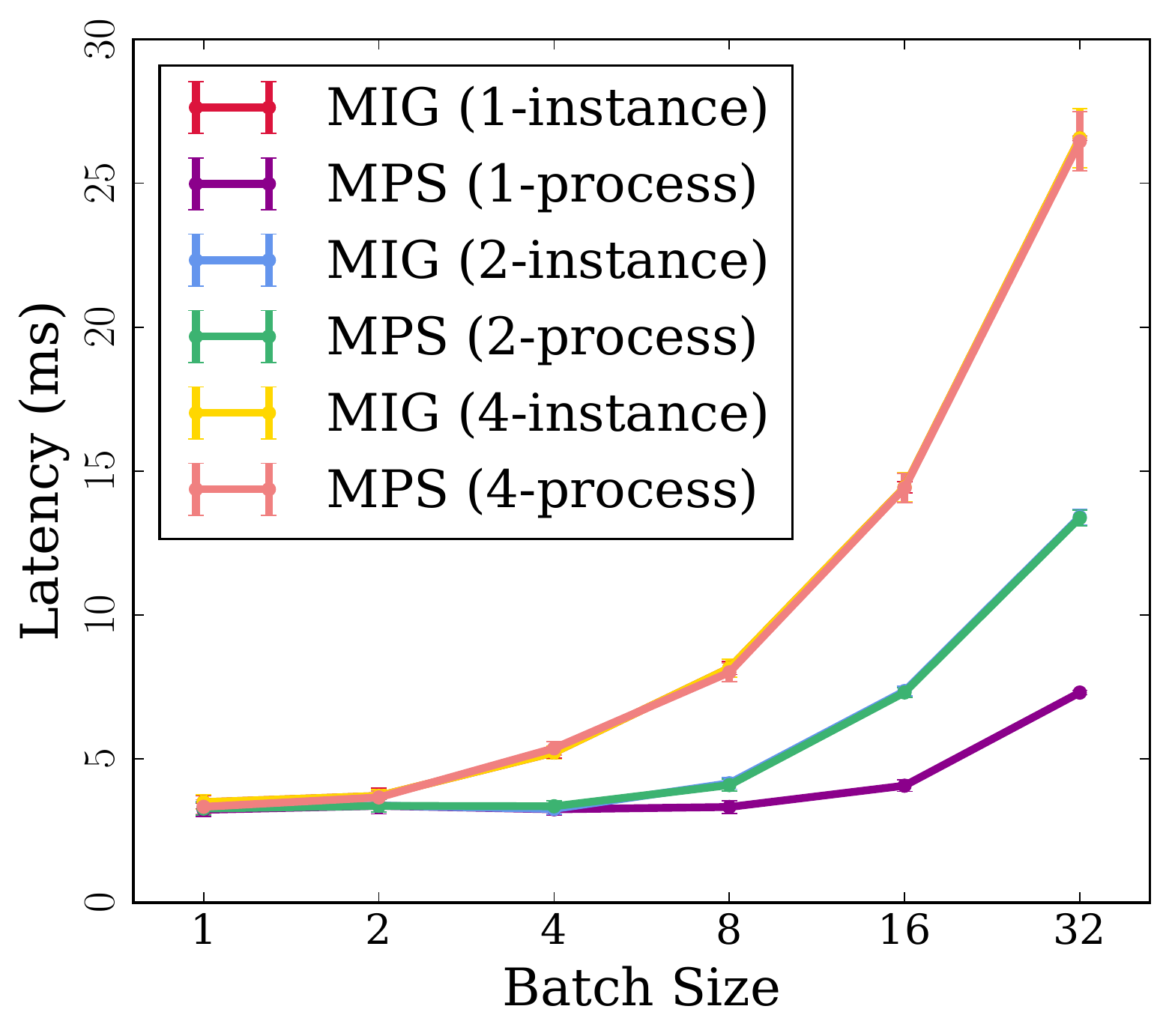}
  \caption{ResNet18 average latency}
  \label{fig:a30_inference_average_latency_mps_mig_resnet18}
\end{subfigure}%
\begin{subfigure}{.5\columnwidth}
  \centering
  \includegraphics[width=1.0\linewidth]{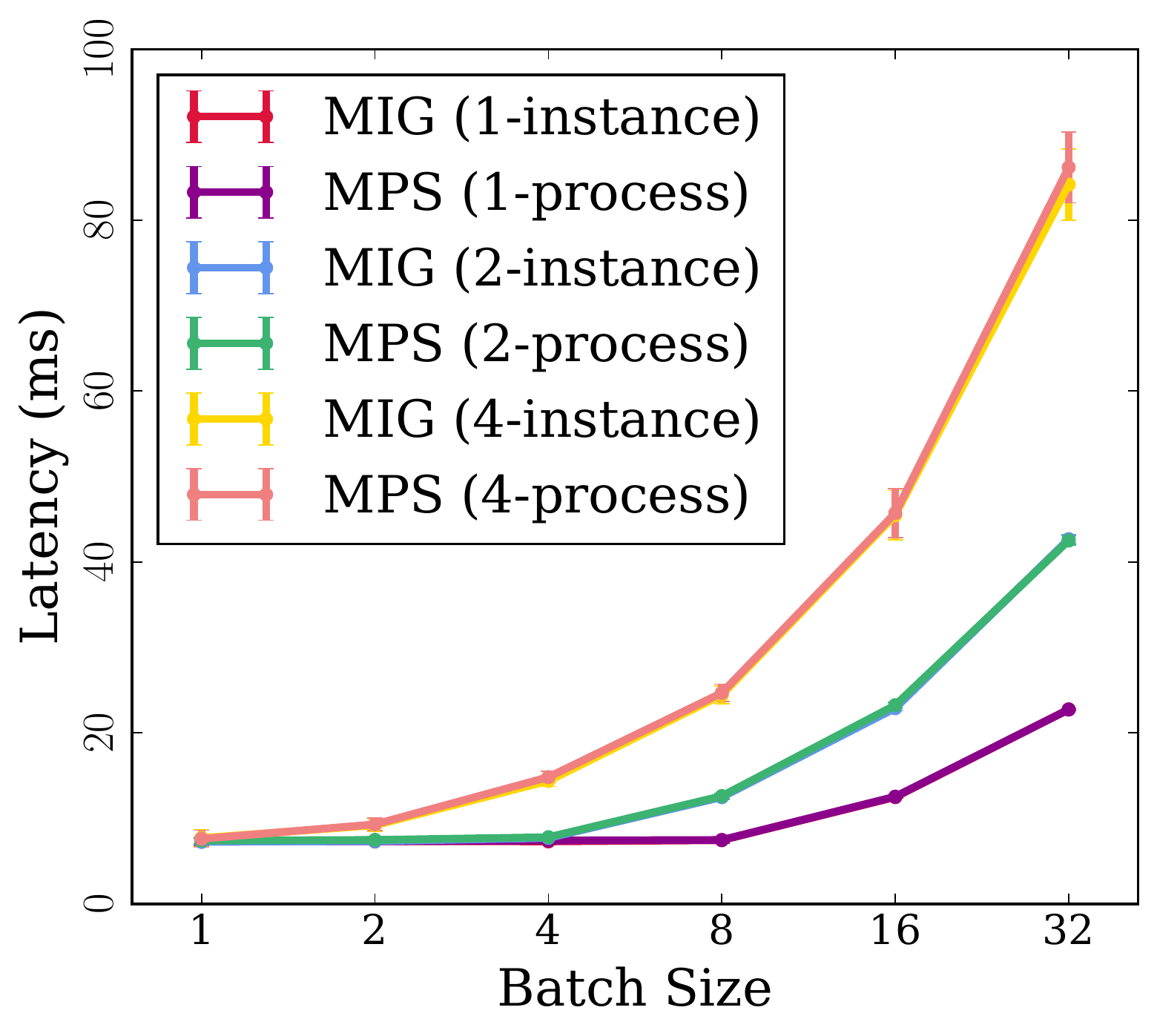}
  \caption{ResNet50 average latency}
  \label{fig:a30_inference_average_latency_mps_mig_resnet50}
\end{subfigure}
\caption{The average latency comparison of different models on MPS and MIG. We present ReNet18 and ResNet-50 results under different batch sizes. The results show that MPS can have a very similar performance to that of MIG when the batch size is small.}
\label{fig:a30_inference_average_latency_mps_mig}
\end{figure}

The first experiment presents the \textbf{average latency comparison}, as shown in Figure \ref{fig:a30_inference_average_latency_mps_mig}. For both two evaluated models, the average latency is almost the same on both MIG and MPS, when the inference batch size is small. As the batch size increases, the standard deviation becomes large, indicating that more interference occurs. This experiment proves that MIG can not always be better than MPS. When a workload is small, even software sharing like MPS can achieve a good performance.

\begin{figure}[t]
\begin{subfigure}{0.5\columnwidth}
  \centering
  \includegraphics[width=1.0\linewidth]{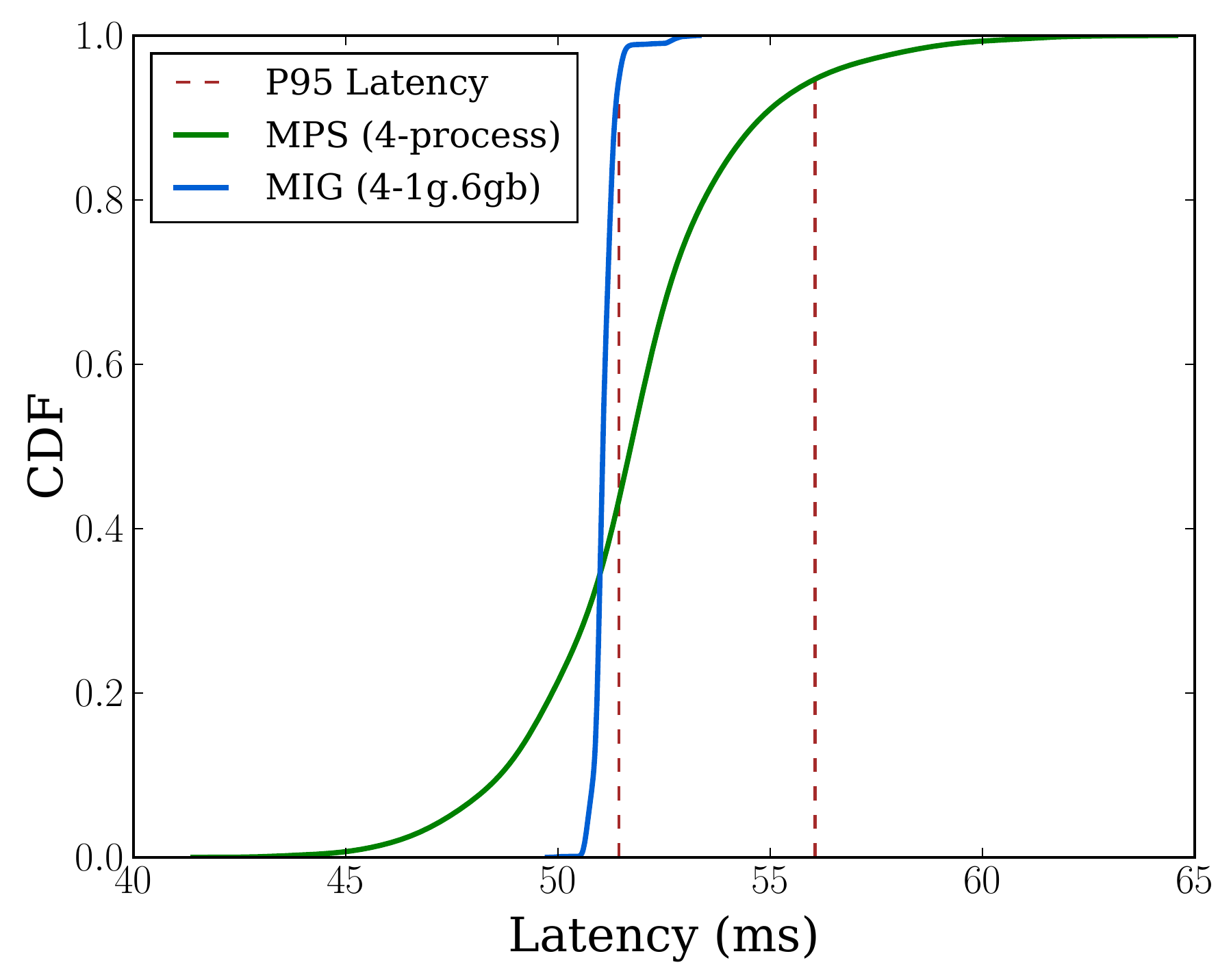}
  \caption{ResNet18 tail latency}
  \label{fig:a30_inference_tail_latency_mps_mig_resnet18}
\end{subfigure}%
\begin{subfigure}{.5\columnwidth}
  \centering
  \includegraphics[width=1.0\linewidth]{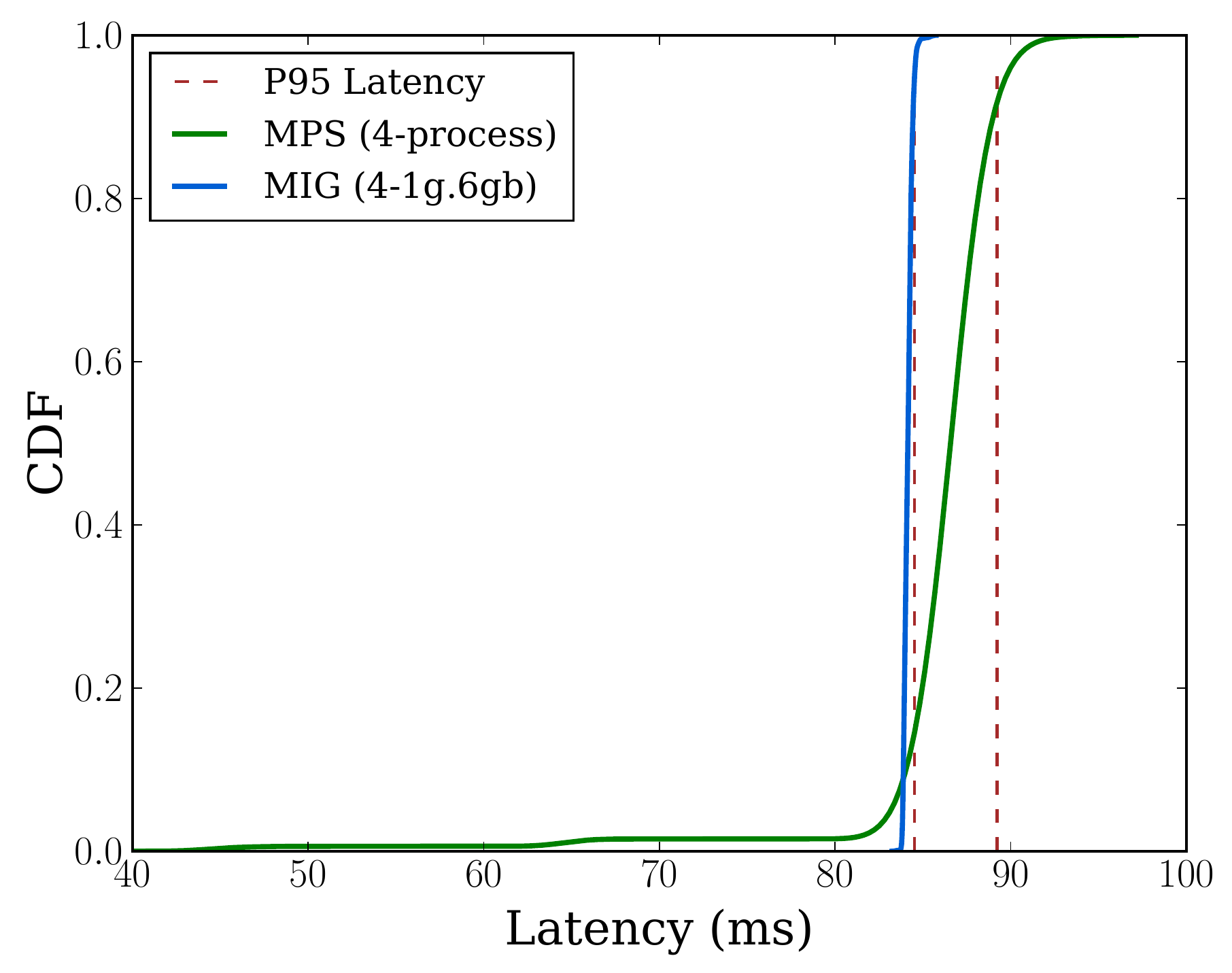}
  \caption{ResNet50 tail latency}
  \label{fig:a30_inference_tail_latency_mps_mig_resnet50}
\end{subfigure}
\caption{The tail latency comparison of different models on MPS and MIG. We set the batch size = 8 and evaluate two models, ResNet18 and ResNet50. The results show that MIG outperforms MPS a lot under the large batch size in terms of both tail latency and stability.}
\label{fig:a30_inference_tail_latency_mps_mig}
\end{figure}

To further explore the occurred interference, we present the \textbf{tail latency comparison}, as shown in Figure \ref{fig:a30_inference_tail_latency_mps_mig}. In this experiment, we set the batch size = 8, as the interference becomes very significant at this point. Figure \ref{fig:a30_inference_tail_latency_mps_mig} shows that from a tail latency perspective, MIG outperforms MPS a lot. MIG has a lower latency and can process users' requests stably. However, if users set a large service level objective (SLO) (for example, in an offline scenario), this advantage will disappear.

\begin{figure}[t]
\begin{subfigure}{0.5\columnwidth}
  \centering
  \includegraphics[width=1.0\linewidth]{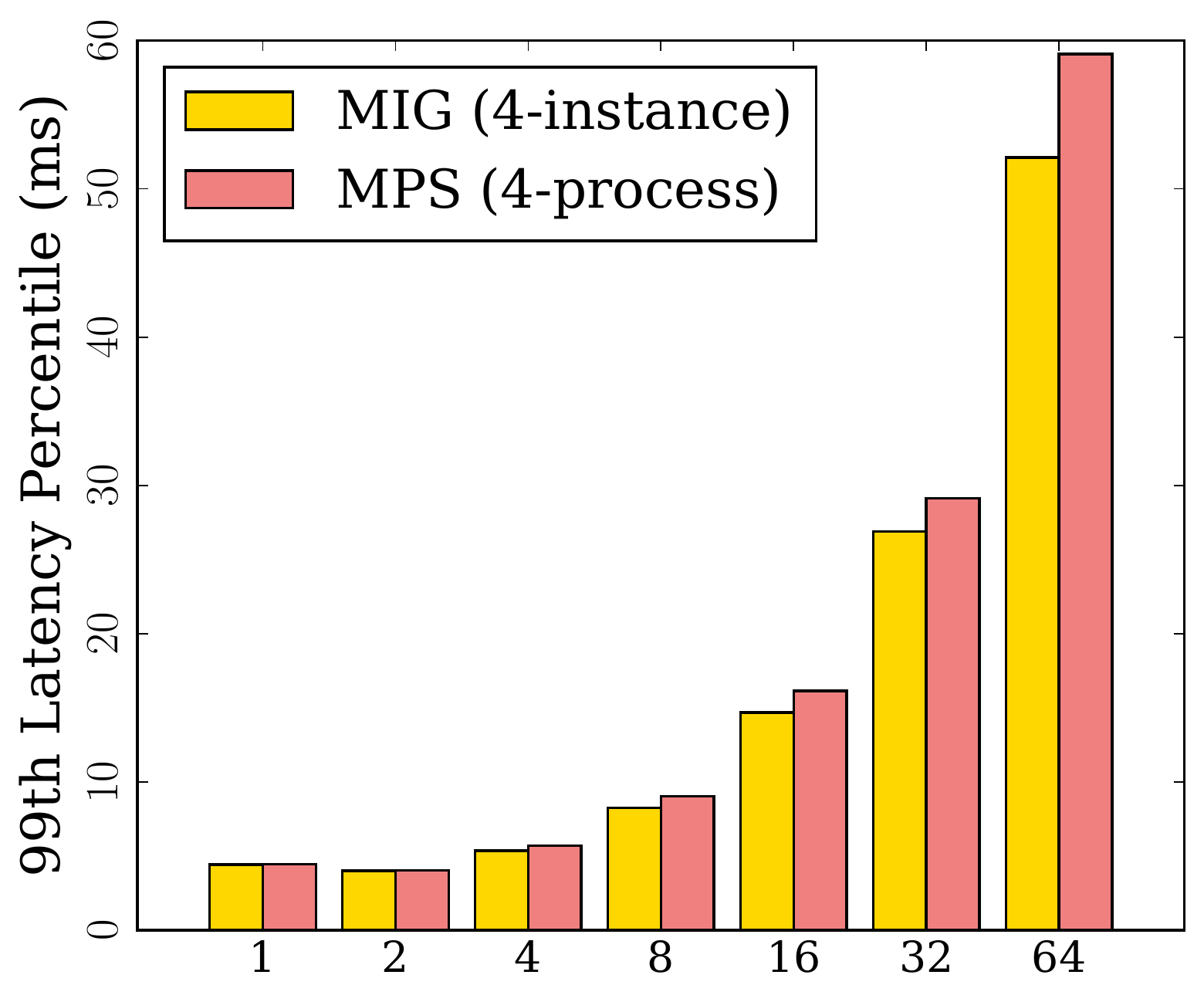}
  \caption{ResNet18 tail latency}
  \label{fig:a30_inference_tail_latency_mps_mig_batch_size_resnet18}
\end{subfigure}%
\begin{subfigure}{.5\columnwidth}
  \centering
  \includegraphics[width=1.0\linewidth]{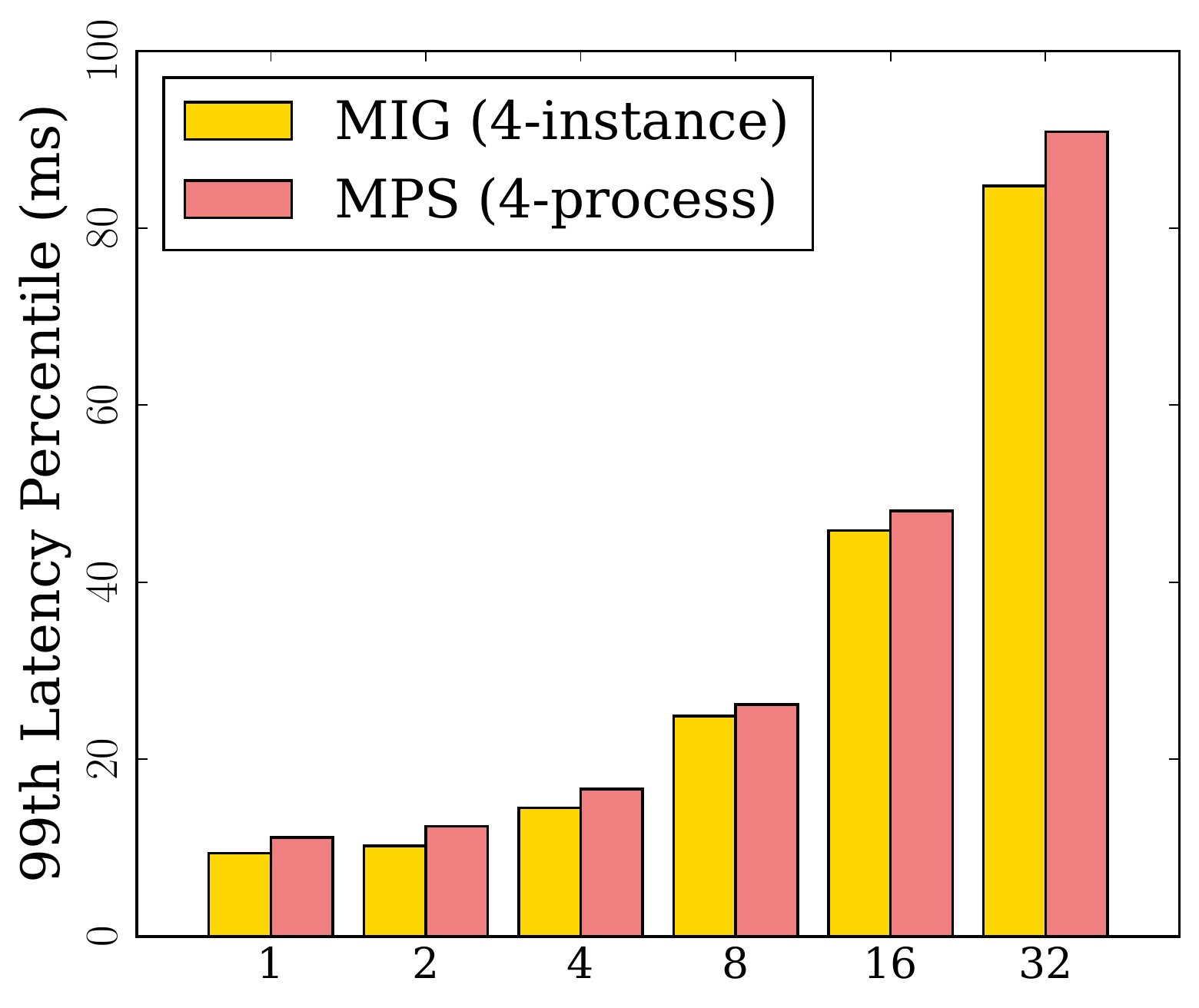}
  \caption{ResNet50 tail latency}
  \label{fig:a30_inference_tail_latency_mps_mig_batch_size_resnet50}
\end{subfigure}
\caption{The tail latency comparison under different batch sizes on MPS and MIG. The experiments show that MPS is comparable to MIG under small batch sizes and will be outperformed with large batch sizes.}
\label{fig:a30_inference_tail_latency_mps_mig_batch_size}
\end{figure}

We further conduct experiments under different batch sizes to verify the above-mentioned conclusion. Figure \ref{fig:a30_inference_tail_latency_mps_mig_batch_size} presents the results. It shows that for both two models on MIG and MPS, the gap of tail latency is very marginal when the batch size is small and becomes larger as the batch size increases. This insight aligns with previous results and demonstrates more details further.

\begin{figure}[t]
\begin{subfigure}{0.5\columnwidth}
  \centering
  \includegraphics[width=1.0\linewidth]{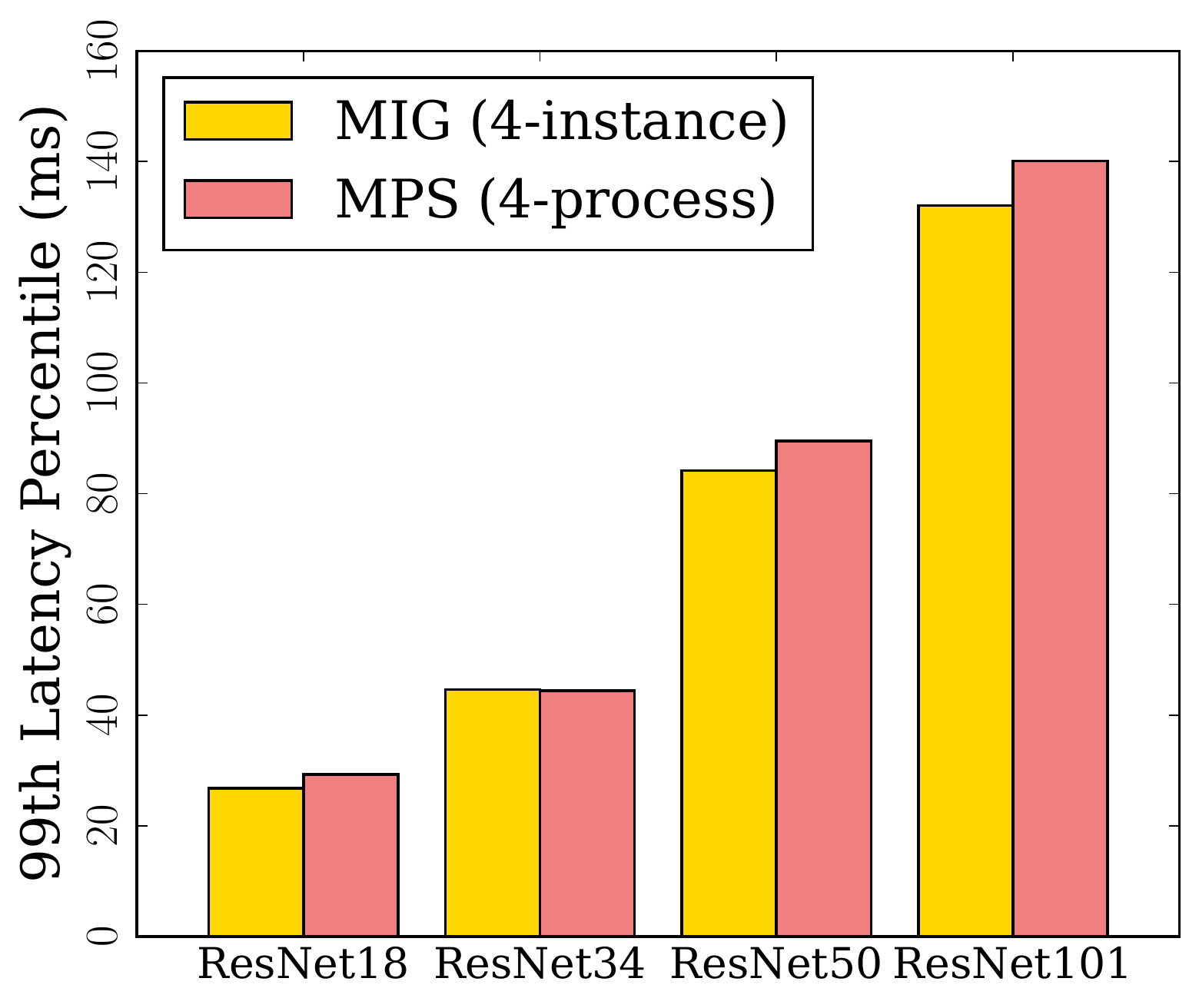}
  \caption{ResNet family models}
  \label{fig:a30_inference_tail_latency_mps_mig_model_resnet}
\end{subfigure}%
\begin{subfigure}{.5\columnwidth}
  \centering
  \includegraphics[width=1.0\linewidth]{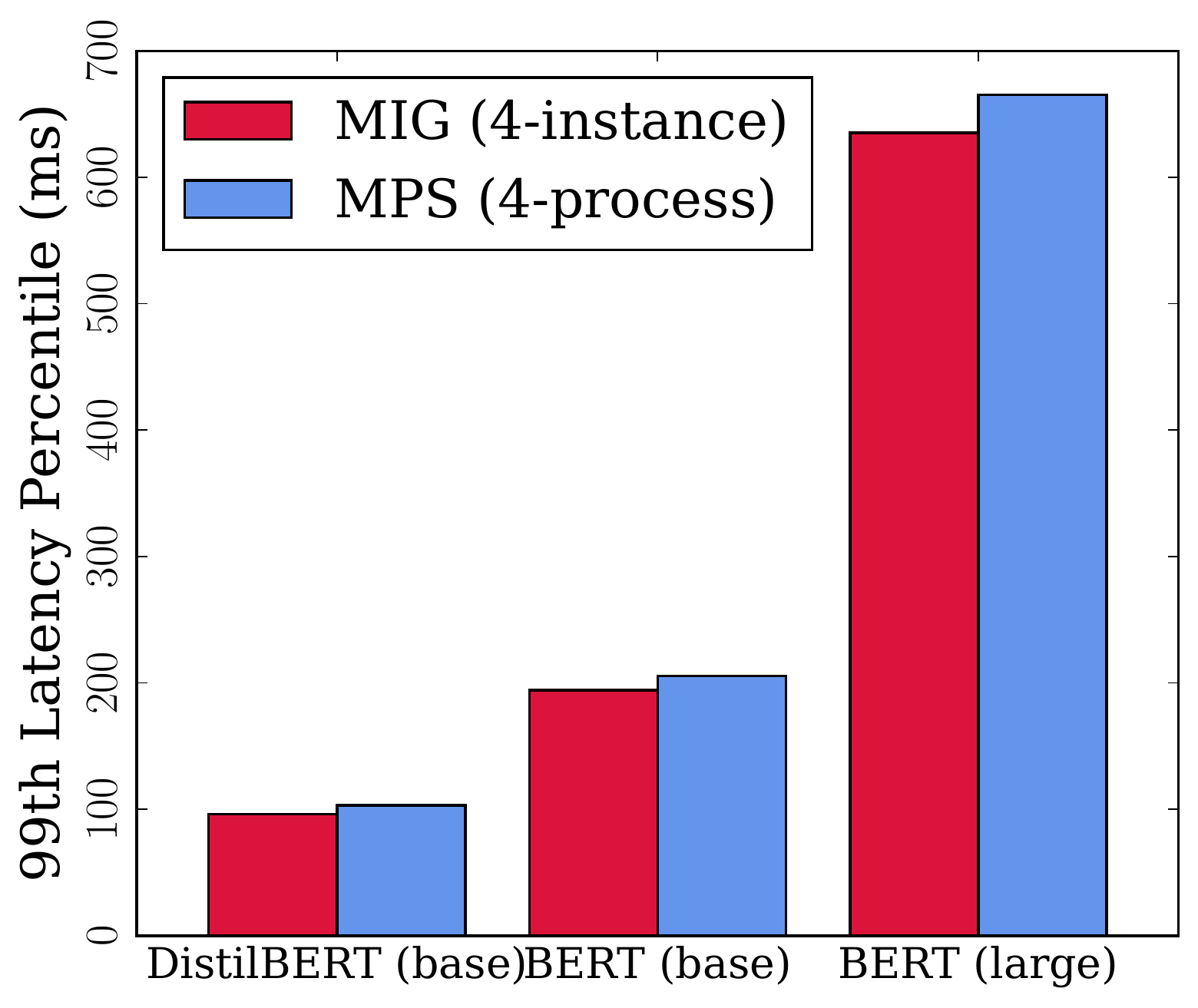}
  \caption{BERT family models}
  \label{fig:a30_inference_tail_latency_mps_mig_model_bert}
\end{subfigure}
\caption{The tail latency comparison of different sized models on MPS and MIG. When the batch size is fixed, both MIG and MPS can support small size models well, but MIG have a lower latency for larger models compared to MPS.}
\label{fig:a30_inference_tail_latency_mps_mig_model}
\end{figure}

The last experiment in this section investigates the performance of different model sizes on MIG and MPS. We set the batch size to 8. Figure \ref{fig:a30_inference_tail_latency_mps_mig_model} both MIG and MPS can handle interference well for small-size models, but MIG will outperform MPS for large-size models. This can be attributed to physical isolation.

\subsection{Framework Compatibility with MIG}
\label{sec:framework_MIG}

In this section, we test DL training and serving frameworks' compatibility with MIG. Unless otherwise stated, all experiments are conducted on A30 GPUs with NVIDIA Driver 470.82.01, CUDA 11.4, and CuDNN 8.2.

Table \ref{tab:training_compat} presents four training frameworks' results. We create two GIs on A30 GPUs. However, due to the limited support from NVIDIA, all tested frameworks can not detect the second GIs. As a result, only the first GI can be used to train. Similarly, for all evaluated serving frameworks in Table \ref{tab:serving_compat}, only MIG 0 can be used to deploy and serve DL models.

This limitation can be initially addressed by utilizing docker techniques. Users can bind one docker container on one GI to perform tasks on one specific GI, which is not limited to MIG 0. However, this also brings other issues - we can not directly divide or adjust GI resources within one docker. That is, every time we want to adjust GIs, we need to stop the running docker containers and unbind them from a GI. Then we reconfigure the GI size, bind a docker to a new GI, and then rerun the docker containers. These complex and unfriendly operations will drive us to propose better methods to improve the training and serving frameworks on MIGs. We leave this a future work.

\begin{table}[]
\caption{Training framework compatibility with MIG. Only the MIG 0 can be used to train for all tested frameworks.}
\label{tab:training_compat}
\resizebox{\columnwidth}{!}{%
\begin{tabular}{@{}lllll@{}}
\toprule
\begin{tabular}[c]{@{}l@{}}Training \\ framework\end{tabular} & Version & \begin{tabular}[c]{@{}l@{}}Visible device \\ count\end{tabular} & \begin{tabular}[c]{@{}l@{}}Training on \\ MIG 0\end{tabular} & \begin{tabular}[c]{@{}l@{}}Training on \\ MIG 1\end{tabular} \\ \midrule
PyTorch & 1.13.0 & 0 & Yes & No device \\
TensorFlow & 2.11.0 & 1 & Yes & No device \\
MxNet & 1.9.1 & 1 & Yes & No device \\
PaddlePaddle & 2.4.1 & 1 & Yes & No device \\ \bottomrule
\end{tabular}%
}
\end{table}

\begin{table}[]
\centering
\caption{Serving framework compatibility with MIG. Only the MIG 0 can be found to serve ML models for all tested frameworks.}
\label{tab:serving_compat}
\begin{tabular}{@{}llll@{}}
\toprule
\begin{tabular}[c]{@{}l@{}}Serving \\ framework\end{tabular} & Version & \begin{tabular}[c]{@{}l@{}}Serving on \\ MIG 0\end{tabular} & \begin{tabular}[c]{@{}l@{}}Serving on \\ MIG 1\end{tabular} \\ \midrule
\begin{tabular}[c]{@{}l@{}}TensorFlow \\ Serving\end{tabular} & 2.8.4 & Yes & \begin{tabular}[c]{@{}l@{}}Device not \\ found\end{tabular} \\
\begin{tabular}[c]{@{}l@{}}Triton Inference \\ Server\end{tabular} & 21.09 & Yes & \begin{tabular}[c]{@{}l@{}}Device not \\ found\end{tabular} \\
Ray Serve & 2.2.0 & Yes & \begin{tabular}[c]{@{}l@{}}Device not \\ found\end{tabular} \\ \bottomrule
\end{tabular}
\end{table}

\section{Summary and Future Work}
\label{sec:summary}

MIG is a technology that allows users to easily orchestrate training and inference workloads for deep learning applications. MIGPerf is an open-source tool that provides a comprehensive overview of MIG by conducting a range of benchmark studies. These studies include training and inference characterization, GPU sharing comparisons, and framework compatibility. The results of these studies provide insight into how MIG can be used effectively and efficiently. In addition, the results point to several promising directions for future research, such as hybrid scheduling for training and inference on MIG and MIG/MPS orchestration.

\newpage
\bibliography{example_paper}
\bibliographystyle{mlsys2020}

\clearpage
\appendix
\section{Detailed Evaluation Settings of the Benchmark Study}
\label{sec:exp_detail}
\textbf{Hardware.} As mentioned in the \ref{eval_setting}, we conduct the benchmark study on two different GPU servers, A30 servers and A100 servers respectively. Table \ref{tab:gpu-server-details} lists the details of these servers.

\begin{table}[ht]
\caption{Hardware details of two GPU servers used for the benchmark study.}
\resizebox{\columnwidth}{!}{%
\begin{tabular}{lll}
\toprule
                      & \textbf{A100 Server}          & \textbf{A30 Server}   \\
\midrule
Physical CPU Model    & Intel Xeon Platinum 8369B     & AMD EPYC 7302P        \\
Number of CPU Socket  & 2                             & 1                     \\
Number of CPU Core    & 64                            & 16                    \\
Number of vCPU        & 128                           & 32                    \\
Memory Size           & 1024 GiB                      & 128 GiB               \\
Memory Channels       & 32                            & 8                     \\
Memory Type           & DDR4 3200 MT/s                & DDR4 3200 MT/s        \\
GPU Model             & 8 $\times$ NVIDIA A100 (80GB) & 2 $\times$ NVIDIA A30 \\
NVIDIA Driver Version & 470.82.01                     & 515.65.01             \\
CUDA Version          & 11.4                          & 11.6                  \\
CuDNN Version         & 8.2                           & N.A.                  \\
OS                    & CentOS 7.9.2009               & Ubuntu 20.04          \\
Linux Kernel          & 3.10.0-1160.80.1.el7 x86\_64   & 5.15.0-56-generic    \\
\bottomrule
\end{tabular}%
}
\label{tab:gpu-server-details}
\end{table}

\textbf{Models.} We have benchmarked MIG training and inference characterization on many open-sourced deep learning models. All model details are in Table \ref{tab:exp-model-details}.

\begin{table}[ht]
\resizebox{\columnwidth}{!}{%
\begin{tabular}{@{}lll@{}}
\toprule
\textbf{ML Task}                              & \textbf{Model Name} & \begin{tabular}[c]{@{}l@{}}\textbf{Open Sourced} \\ \textbf{Model Repository}\end{tabular} \\ 
\midrule
\multirow{4}{*}{Image Classification}         & ResNet-18           & Torch Hub                              \\
                                              & ResNet-34           & Torch Hub                              \\
                                              & ResNet-50           & Torch Hub                              \\
                                              & ResNet-101          & Torch Hub                              \\
\midrule
\multirow{3}{*}{Text Sequence Classification} & Distil BERT         & Hugging Face                           \\
                                              & BERT                & Hugging Face                           \\
                                              & BERT Large          & Hugging Face                           \\                                              
\bottomrule 
\end{tabular}%
}
\caption{Details of used deep learning models in this paper}
\label{tab:exp-model-details}
\end{table}

\section{MIG Training and Inference Characterization}
\label{Appendix_results}
\begin{figure*}[!b]
\begin{subfigure}{0.5\columnwidth}
  \centering
  \includegraphics[width=1.0\linewidth]{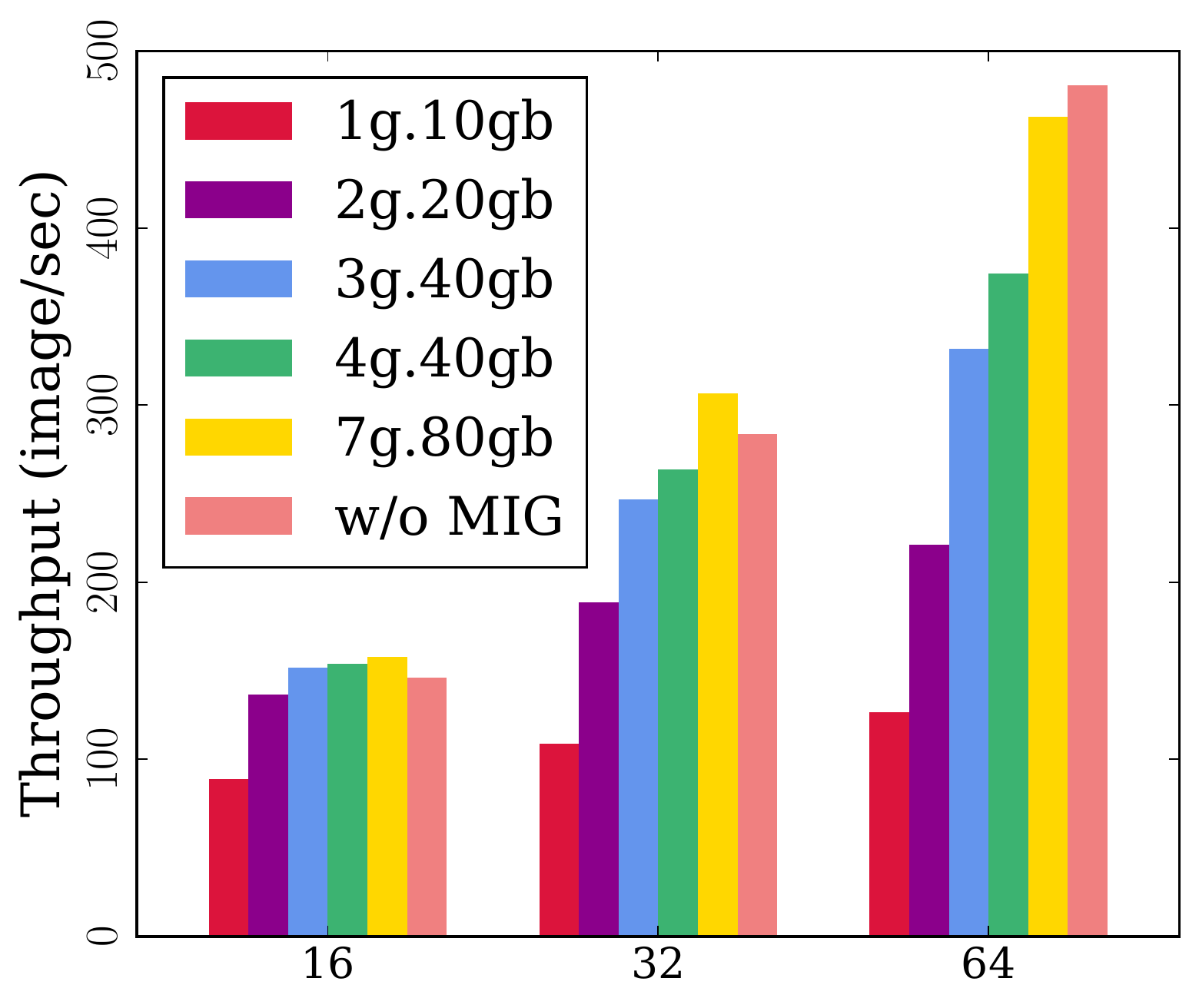}
  \caption{Throughput (batch/sec)}
  \label{fig:single_training_resnet_qps}
\end{subfigure}%
\begin{subfigure}{.5\columnwidth}
  \centering
  \includegraphics[width=1.0\linewidth]{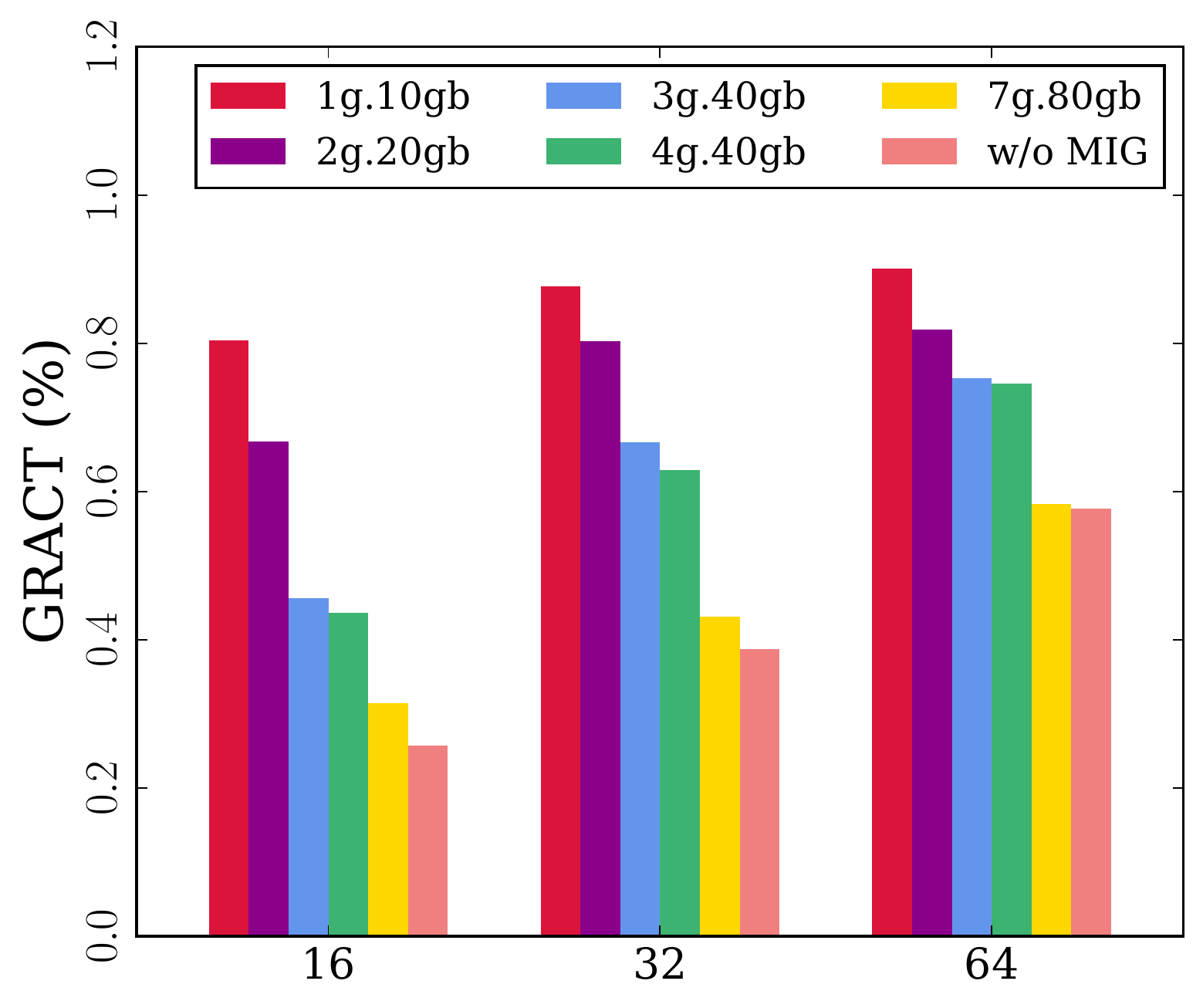}
  \caption{GRACT (\%)}
  \label{fig:single_training_resnet_gract}
\end{subfigure}%
\begin{subfigure}{0.5\columnwidth}
  \centering
  \includegraphics[width=1.0\linewidth]{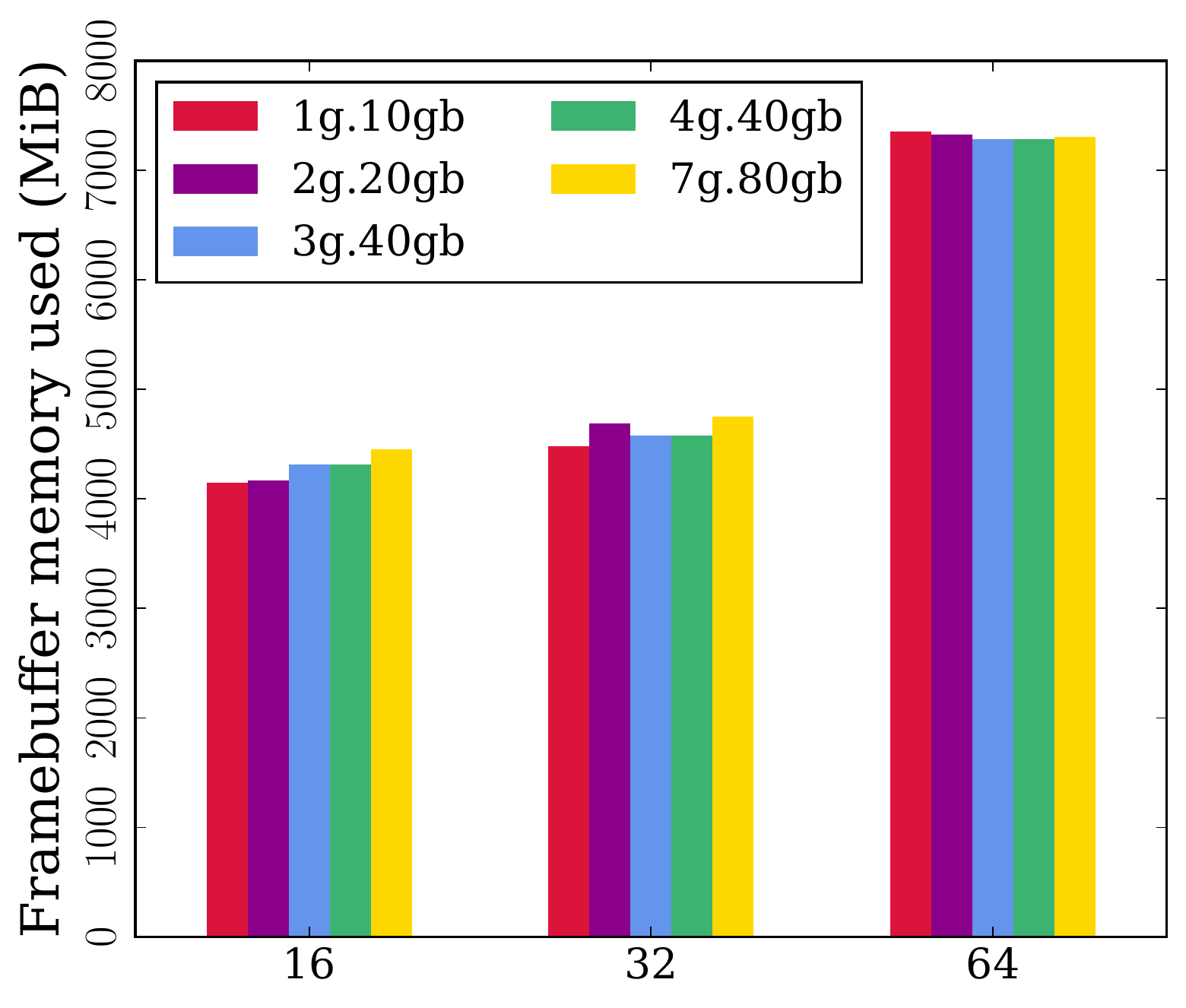}
  \caption{Framebuffer Memory (MiB)}
  \label{fig:single_training_resnet_fb_memory}
\end{subfigure}%
\begin{subfigure}{.5\columnwidth}
  \centering
  \includegraphics[width=1.0\linewidth]{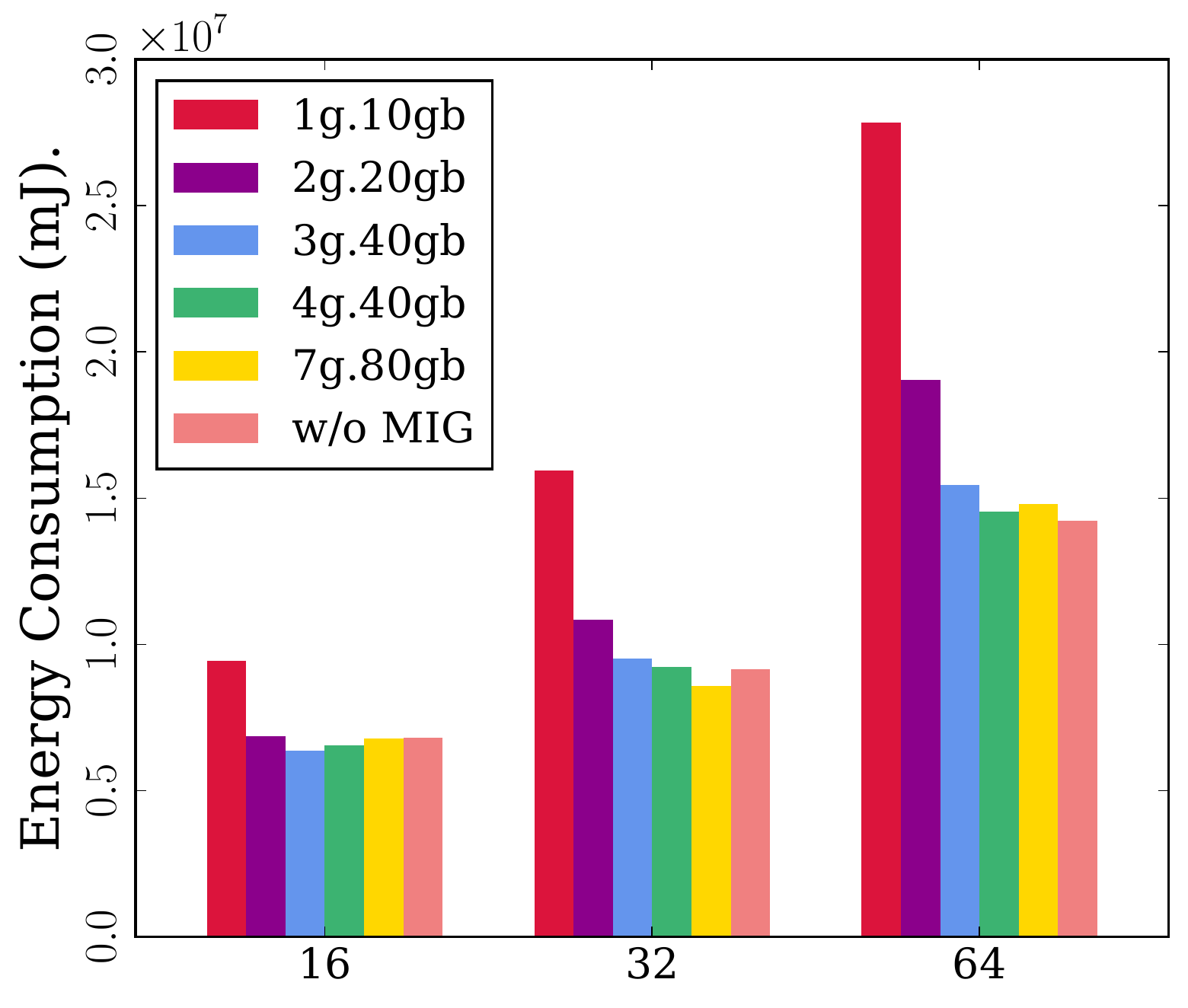}
  \caption{Energy Consumption (mJ)}
  \label{fig:single_training_resnet_bs_energy}
\end{subfigure}
\caption{The impact of input batch size length for training ResNet-50 on single GPU instance (GI) from A100. We present the batch size's influence on throughput, computation utilization, memory utilization, and energy consumption.}
\label{fig:single_training_resnet_batch_size}
\end{figure*}

\begin{figure*}[!b]
\begin{subfigure}{0.5\columnwidth}
  \centering
  \includegraphics[width=1.0\linewidth]{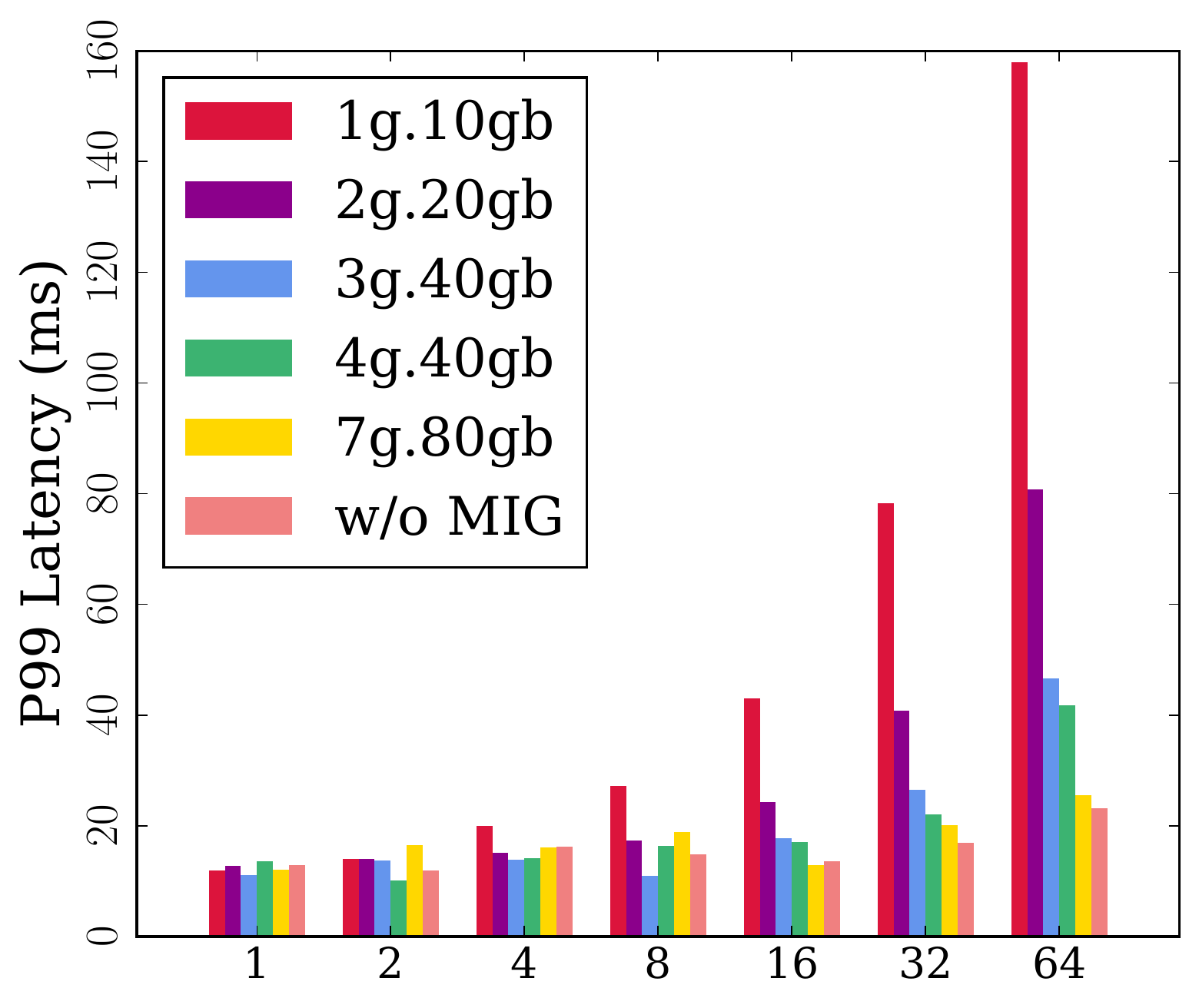}
  \caption{P99 Latency (ms)}
  \label{fig:single_inference_resnet_p99_latency}
\end{subfigure}%
\begin{subfigure}{.5\columnwidth}
  \centering
  \includegraphics[width=1.0\linewidth]{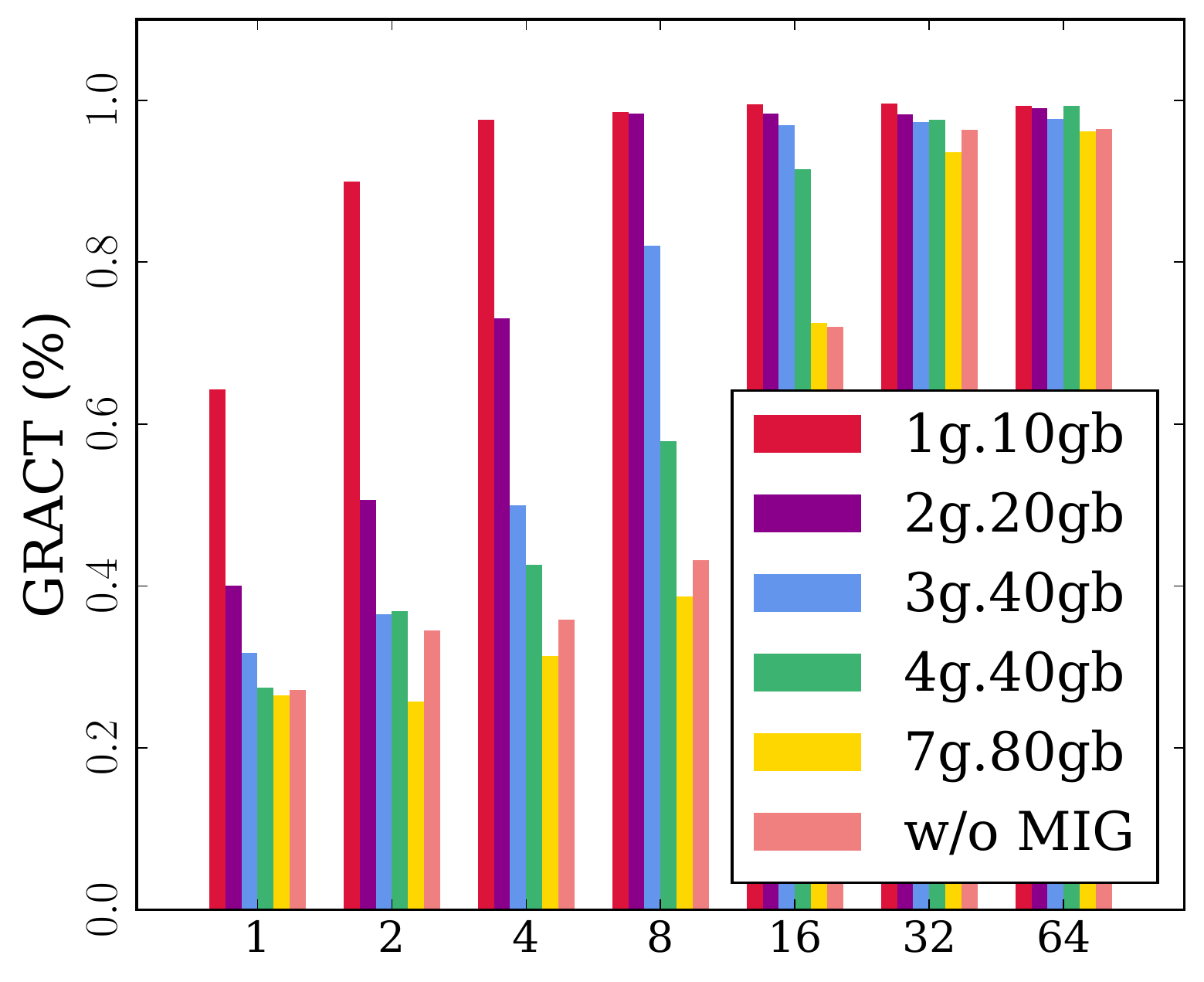}
  \caption{GRACT (\%)}
  \label{fig:single_inference_resnet_utilization}
\end{subfigure}%
\begin{subfigure}{0.5\columnwidth}
  \centering
  \includegraphics[width=1.0\linewidth]{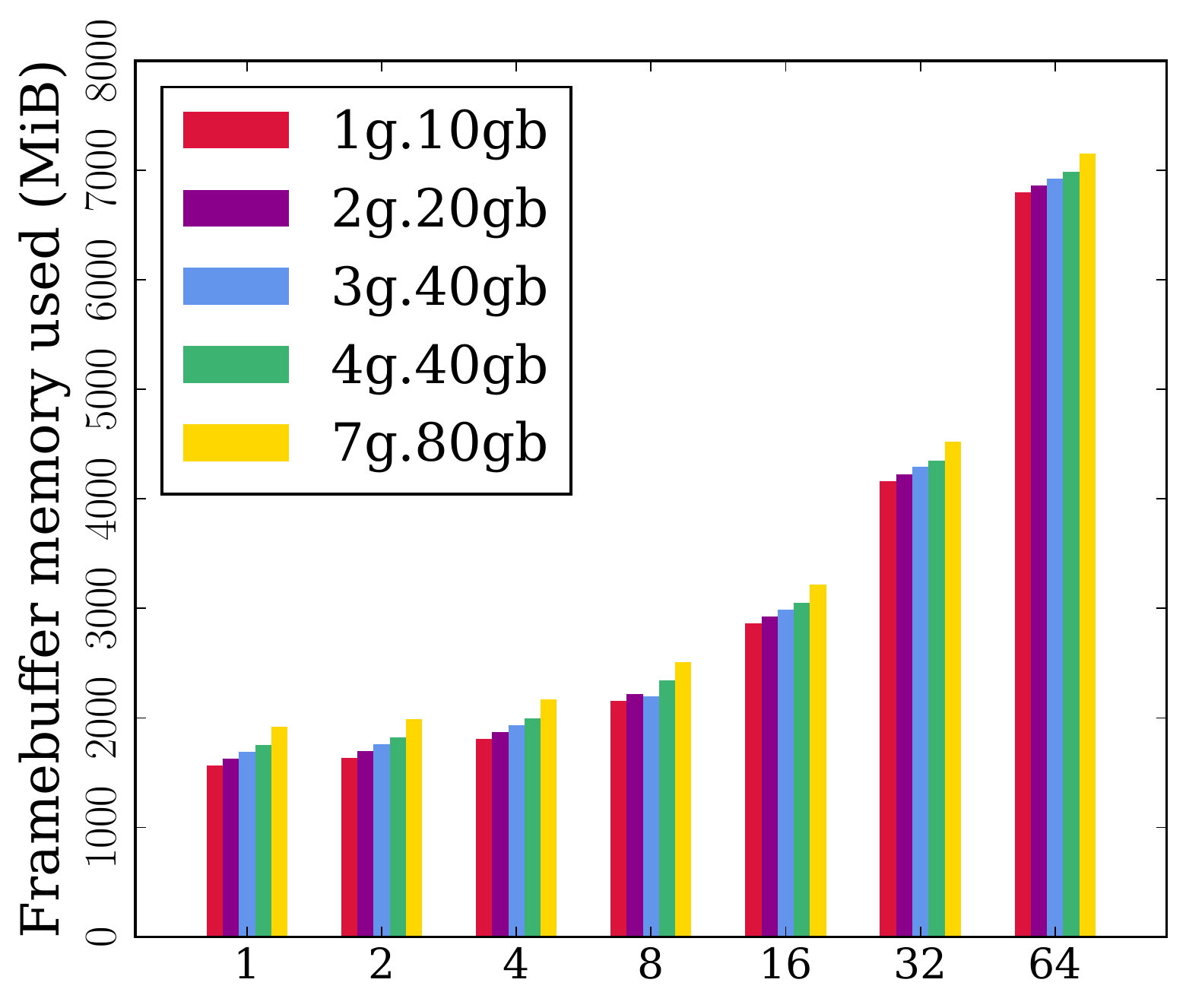}
  \caption{Framebuffer Memory (MiB)}
  \label{fig:single_infeence_resnet_fb}
\end{subfigure}%
\begin{subfigure}{.5\columnwidth}
  \centering
  \includegraphics[width=1.0\linewidth]{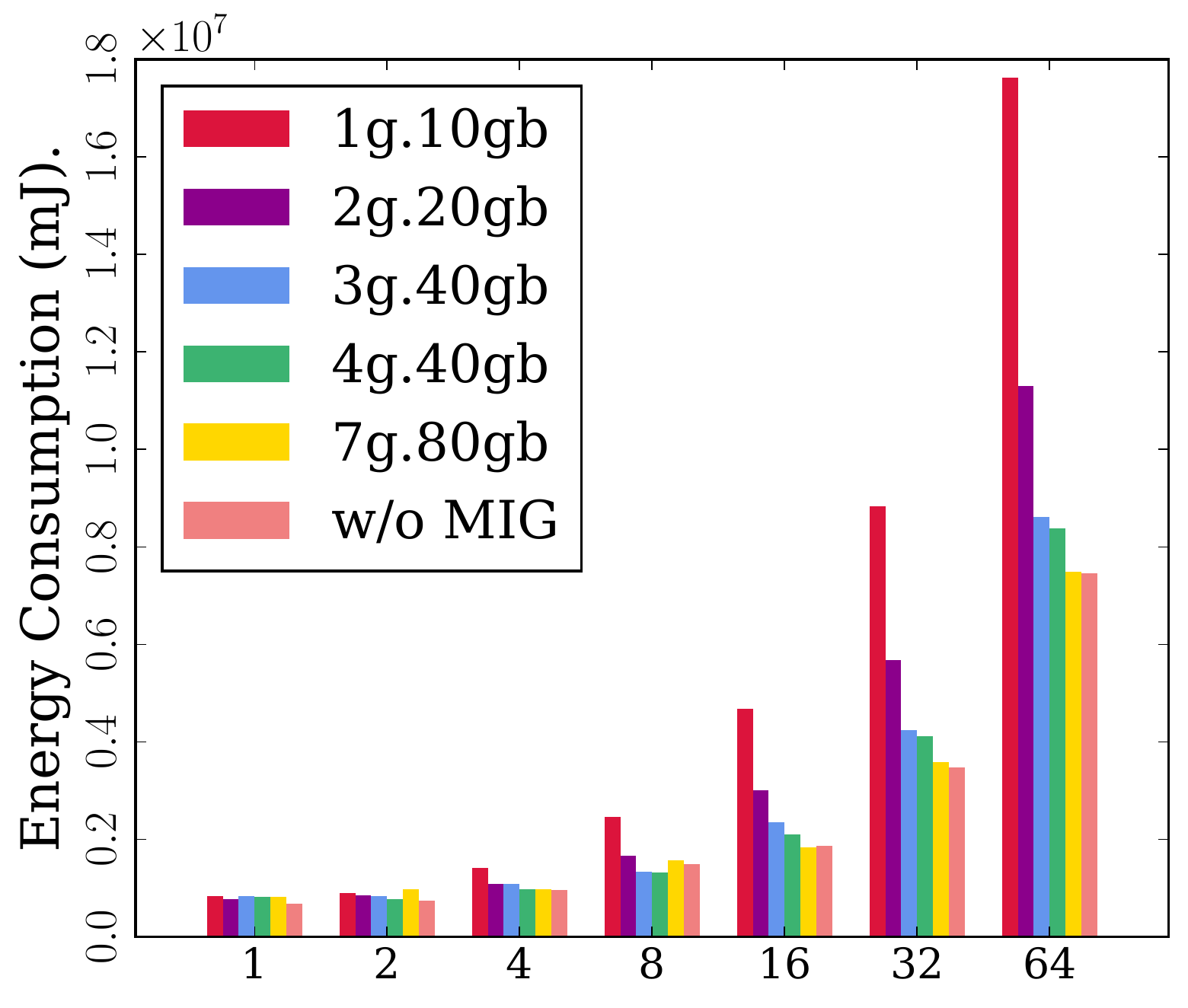}
  \caption{Energy Consumption (mJ)}
  \label{fig:single_inference_resnet_energy}
\end{subfigure}
\caption{The impact of input batch size for ResNet-50 inference on single GPU instance (GI) from A100.}
\label{fig:single_inference_resnet_bs}
\end{figure*}

As mentioned in the previous section \ref{sec:mig_training_exp} and \ref{sec:ming_inference_exp}, we present MIG training and inference benchmark on more open-sourced models. This section shows the ResNet-50's training results in Figure \ref{fig:single_training_resnet_batch_size} and ResNet-50's inference results in Figure \ref{fig:single_inference_resnet_bs}.

\section{GPU Sharing Characterization}

\begin{figure*}[t]
\centering
\begin{subfigure}{0.25\textwidth}
  \centering
  \includegraphics[width=1.0\linewidth]{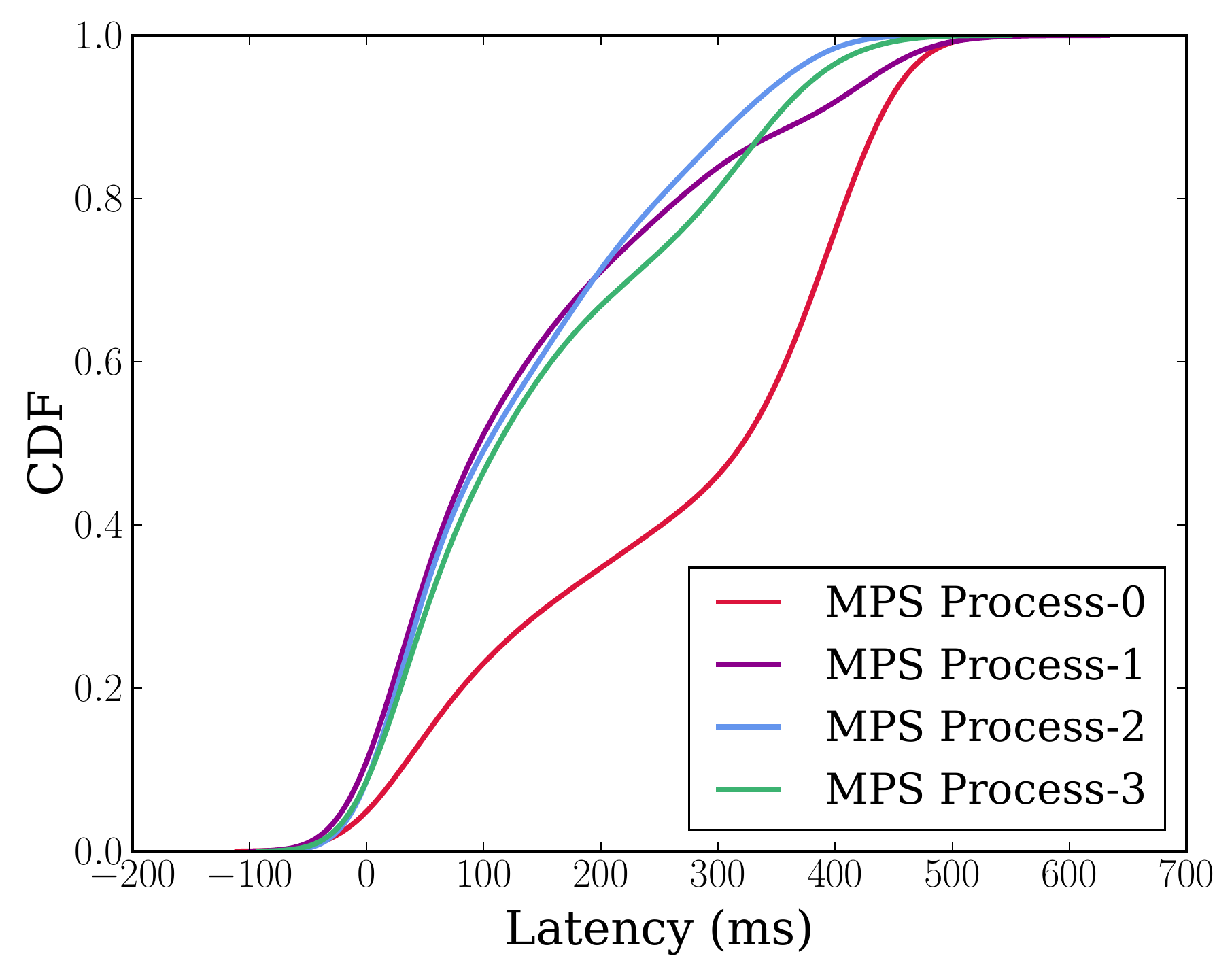}
  \caption{Arrival Rate = 25}
  \label{fig:mps_inference_resnet_arrival_rate25}
\end{subfigure}%
\begin{subfigure}{.25\textwidth}
  \centering
  \includegraphics[width=1.0\linewidth]{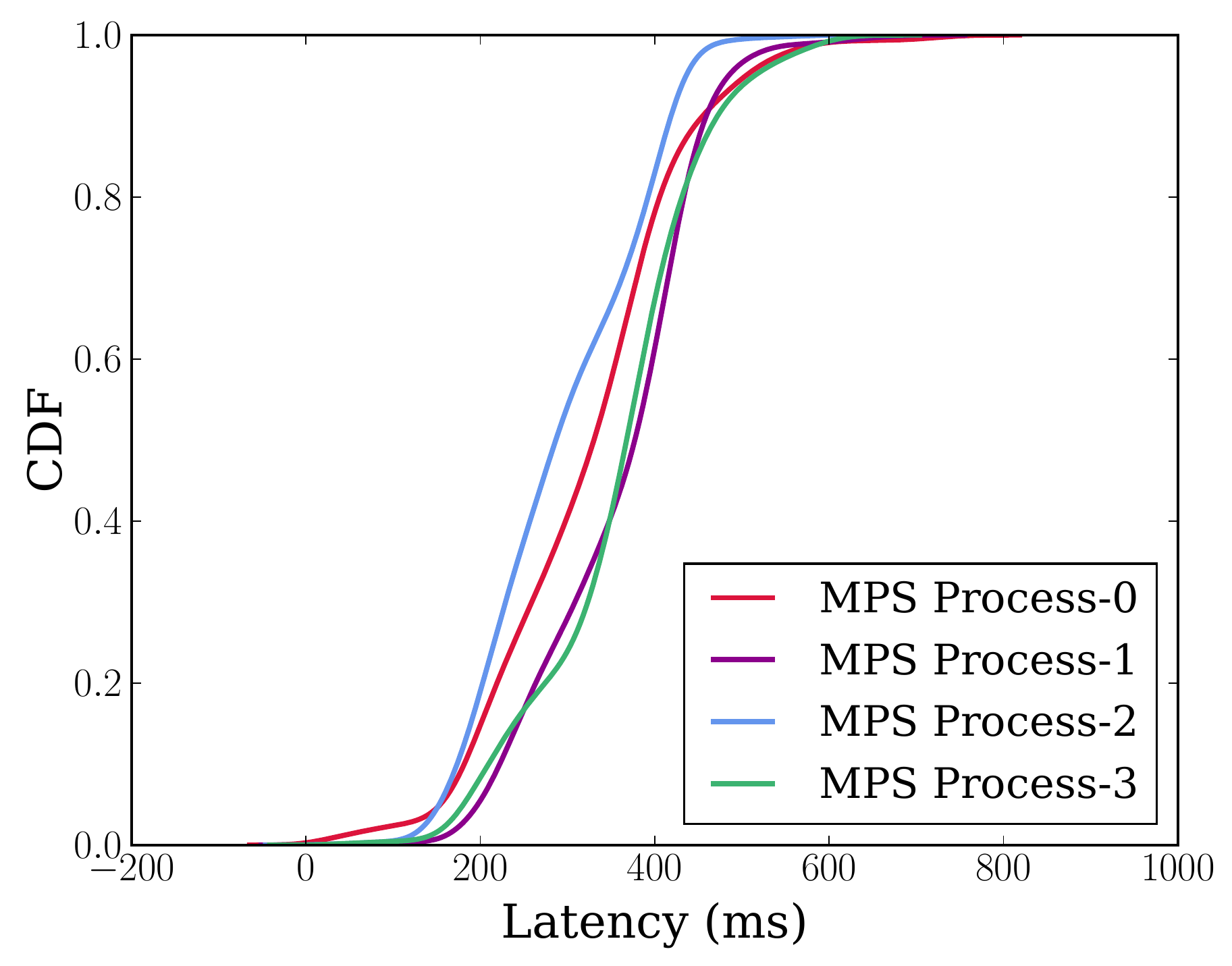}
  \caption{Arrival Rate = 50}
  \label{fig:mps_inference_resnet_arrival_rate50}
\end{subfigure}%
\begin{subfigure}{0.25\textwidth}
  \centering
  \includegraphics[width=1.0\linewidth]{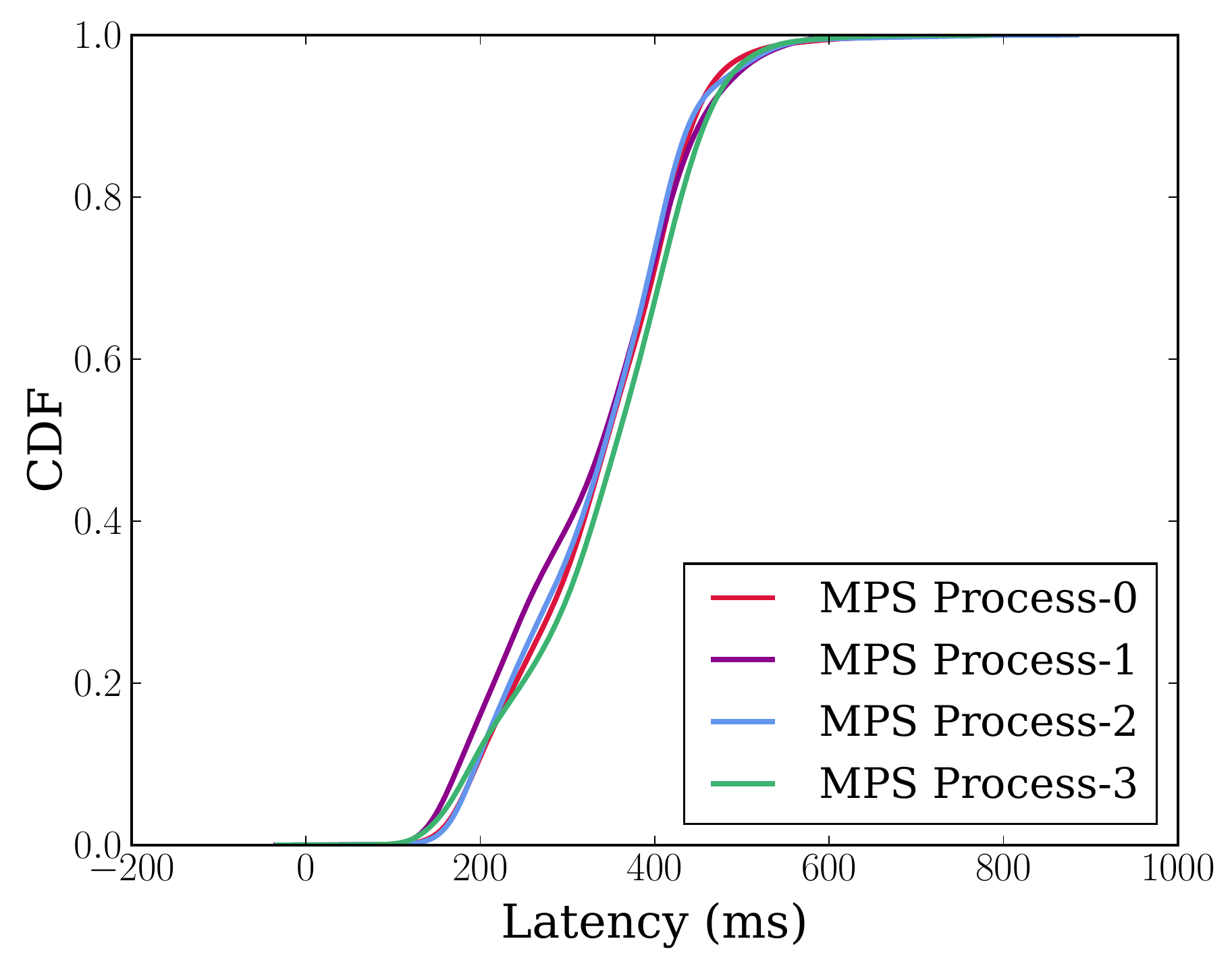}
  \caption{Arrival Rate = 200}
  \label{fig:mps_inference_resnet_arrival_rate200}
\end{subfigure}%
\caption{The tail latency comparison of different request arrival rates for 4 MPS ResNet-50 inference processes on A30.}
\label{fig:mps_inference_resnet_arrival_rate}
\end{figure*}

\begin{figure*}[h]
\centering
\begin{subfigure}{0.25\textwidth}
  \centering
  \includegraphics[width=1.0\linewidth]{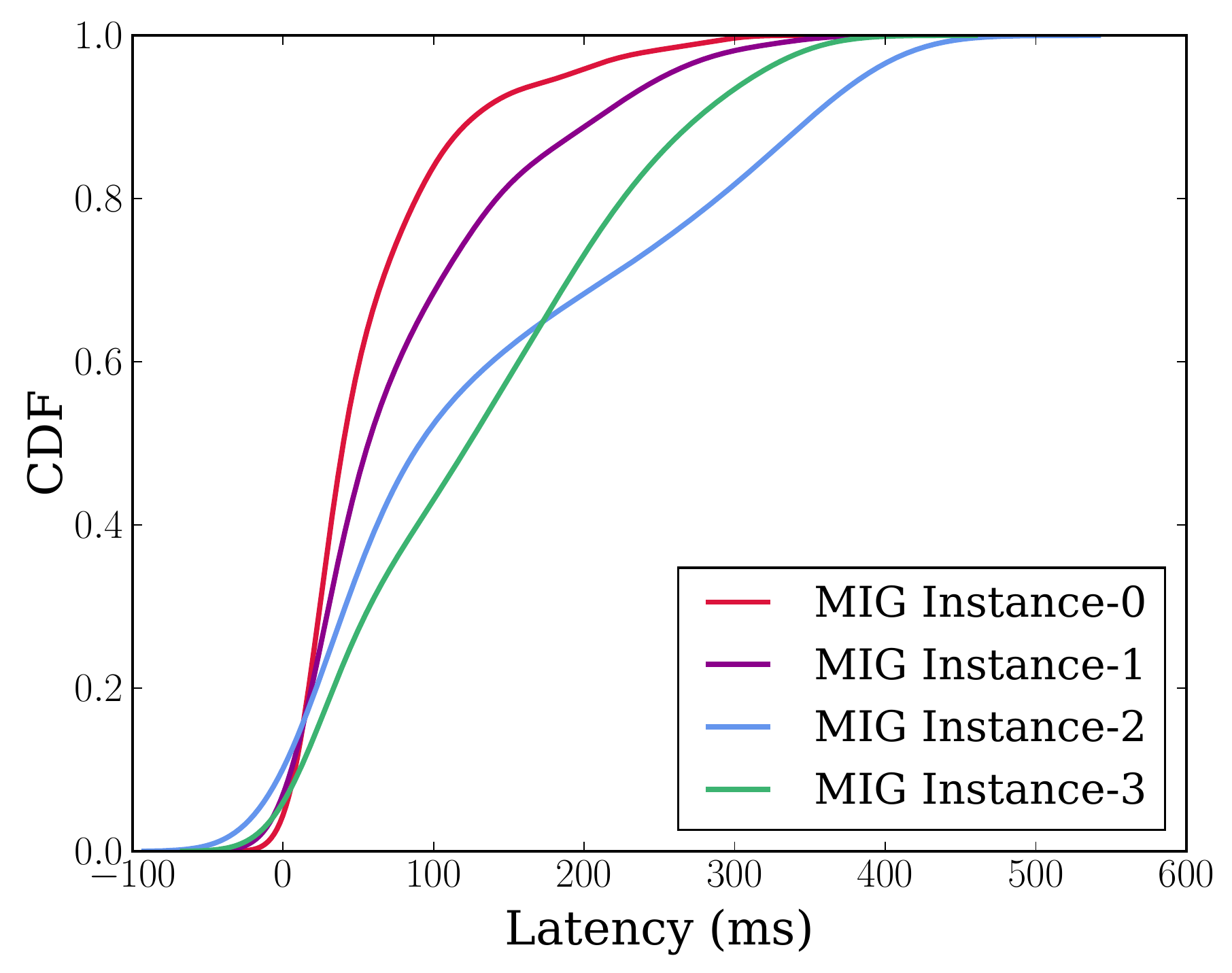}
  \caption{Arrival Rate = 25}
  \label{fig:mig_inference_resnet_arrival_rate25}
\end{subfigure}%
\begin{subfigure}{.25\textwidth}
  \centering
  \includegraphics[width=1.0\linewidth]{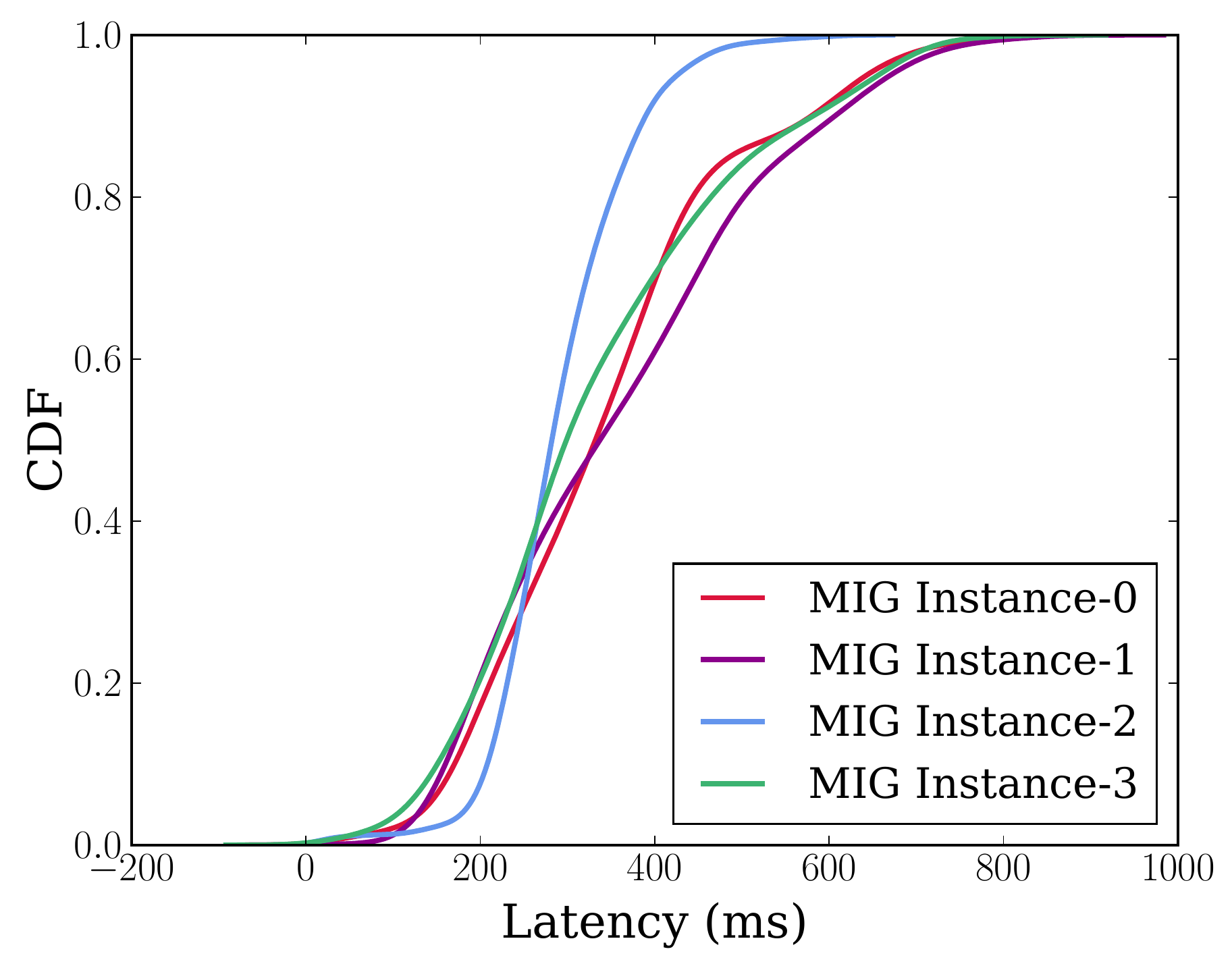}
  \caption{Arrival Rate = 50}
  \label{fig:mig_inference_resnet_arrival_rate50}
\end{subfigure}%
\begin{subfigure}{0.25\textwidth}
  \centering
  \includegraphics[width=1.0\linewidth]{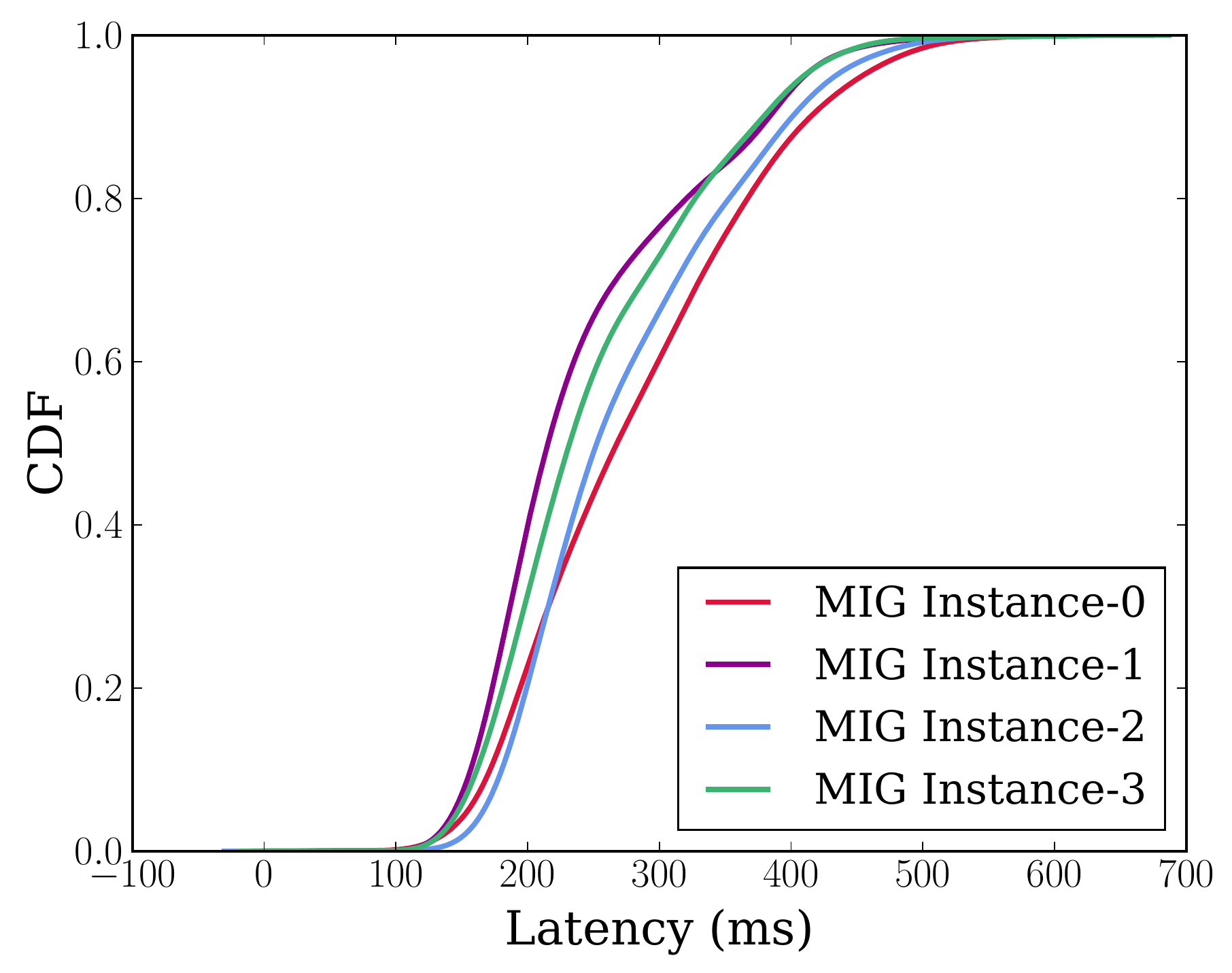}
  \caption{Arrival Rate = 200}
  \label{fig:mig_inference_resnet_arrival_rate200}
\end{subfigure}%
\caption{The tail latency comparison of different request arrival rates for 4 ResNet-50 inference processes on 4 MIG 1g.6gb GPU instances from A30.}
\label{fig:mig_inference_resnet_arrival_rate}
\end{figure*}

In addition to the experiments in section \ref{sec:mig_mps_exp}, we investigate the performance under different online inference request workloads on MIG and MPS. We run 4 simple PyTorch inference servers, and send asynchronous requests to each server simultaneously with different request workloads (i.e., request arrival rate). We set the batch size = 1. Figure \ref{fig:mps_inference_resnet_arrival_rate} and Figure \ref{fig:mig_inference_resnet_arrival_rate} shows the tail latency of MPS and MIG under different online request workload.

\end{document}